\documentclass[journal]{IEEEtran}

\usepackage{xcolor,soul,framed} 

\colorlet{shadecolor}{yellow}
\usepackage[pdftex]{graphicx}
\graphicspath{{../pdf/}{../jpeg/}}
\DeclareGraphicsExtensions{.pdf,.jpeg,.png}

\usepackage[cmex10]{amsmath}
\usepackage{array}
\usepackage{mdwmath}
\usepackage{mdwtab}
\usepackage{eqparbox}
\usepackage{url}

\usepackage{amssymb}
\usepackage{bm}

\usepackage{cite}
\usepackage[subrefformat=parens]{subcaption}

\newcommand{\Tref}[1]{Table~\ref{#1}}
\newcommand{\Eref}[1]{Eq.~(\ref{#1})}

\newcommand{\Fref}[1]{Fig.~\ref{#1}}
\newcommand{\Sref}[1]{Sec.~\ref{#1}}

\newcommand{\MP}[1]{\rm {MP}_{#1}}
\newcommand{\COLEMD}[1]{\rm {Col\mathchar`-EMD}_{#1}}

\newcommand{\INDEMD}[1]{\rm {Ind\mathchar`-EMD}_{#1}}
\newcommand{\INDEMDstr}{Ind-EMD}
\newcommand{\EMDc}{\rm EMD_c}
\newcommand{\tabcenter}[1]{\multicolumn{1}{c}{#1}}

\newcommand{\etal}{\textit{et~al}.}
\newcommand{\etalp}{\textit{et~al}}

\begin{document}
\bstctlcite{IEEEexample:BSTcontrol}
    \title{Image Aesthetics Prediction Using Multiple Patches Preserving the Original Aspect Ratio of Contents}
  \author{Lijie~Wang,~\IEEEmembership{Student Member,~IEEE,}
      Xueting~Wang,~\IEEEmembership{Member,~IEEE,}
      and Toshihiko~Yamasaki,~\IEEEmembership{Member,~IEEE.} \\
      
  \thanks{L. Wang is with the Department of Computer Science, Graduate School of Information Science and Technology, The University of Tokyo, Tokyo 113-8656, Japan (e-mail: wang@hal.t.u-tokyo.ac.jp).}
  \thanks{X. Wang and T. Yamasaki are with the Department of Information and Communication Engineering, Graduate School of Information Science and Technology, The University of Tokyo, Tokyo 113-8656, Japan (e-mail: xt\_wang@hal.t.u-tokyo.ac.jp; yamasaki@hal.t.u-tokyo.ac.jp.}
  \thanks{This is an extended and revised version of a conference paper presented at CVPRW 2019~\cite{wang2019mpemd}. Our source code will be made available if the paper is accepted.}
}


\maketitle

\begin{abstract}
The spread of social networking services has created an increasing demand for selecting, editing, and generating impressive images. This trend increases the importance of evaluating image aesthetics as a complementary function of automatic image processing. We propose a multi-patch method, named MPA-Net (Multi-Patch Aggregation Network), to predict image aesthetics scores by maintaining the original aspect ratios of contents in the images. Through an experiment involving the large-scale AVA dataset, which contains 250,000 images, we show that the effectiveness of the equal-interval multi-patch selection approach for aesthetics score prediction is significant compared to the single-patch prediction and random patch selection approaches. For this dataset, MPA-Net outperforms the neural image assessment algorithm, which was regarded as a baseline method. In particular, MPA-Net yields a 0.073 (11.5\%) higher linear correlation coefficient (LCC) of aesthetics scores and a 0.088 (14.4\%) higher Spearman's rank correlation coefficient (SRCC). MPA-Net also reduces the mean square error (MSE) by 0.0115 (4.18\%) and achieves results for the LCC and SRCC that are comparable to those of the state-of-the-art continuous aesthetics score prediction methods. Most notably, MPA-Net yields a significant lower MSE especially for images with aspect ratios far from 1.0, indicating that MPA-Net is useful for a wide range of image aspect ratios. MPA-Net uses only images and does not require external information during the training nor prediction stages. Therefore, MPA-Net has great potential for applications aside from aesthetics score prediction such as other human subjectivity prediction.

\end{abstract}


%
\IEEEpeerreviewmaketitle


\begin{figure*}[bt]
  \centering
  \scalebox{0.7}{
  \includegraphics{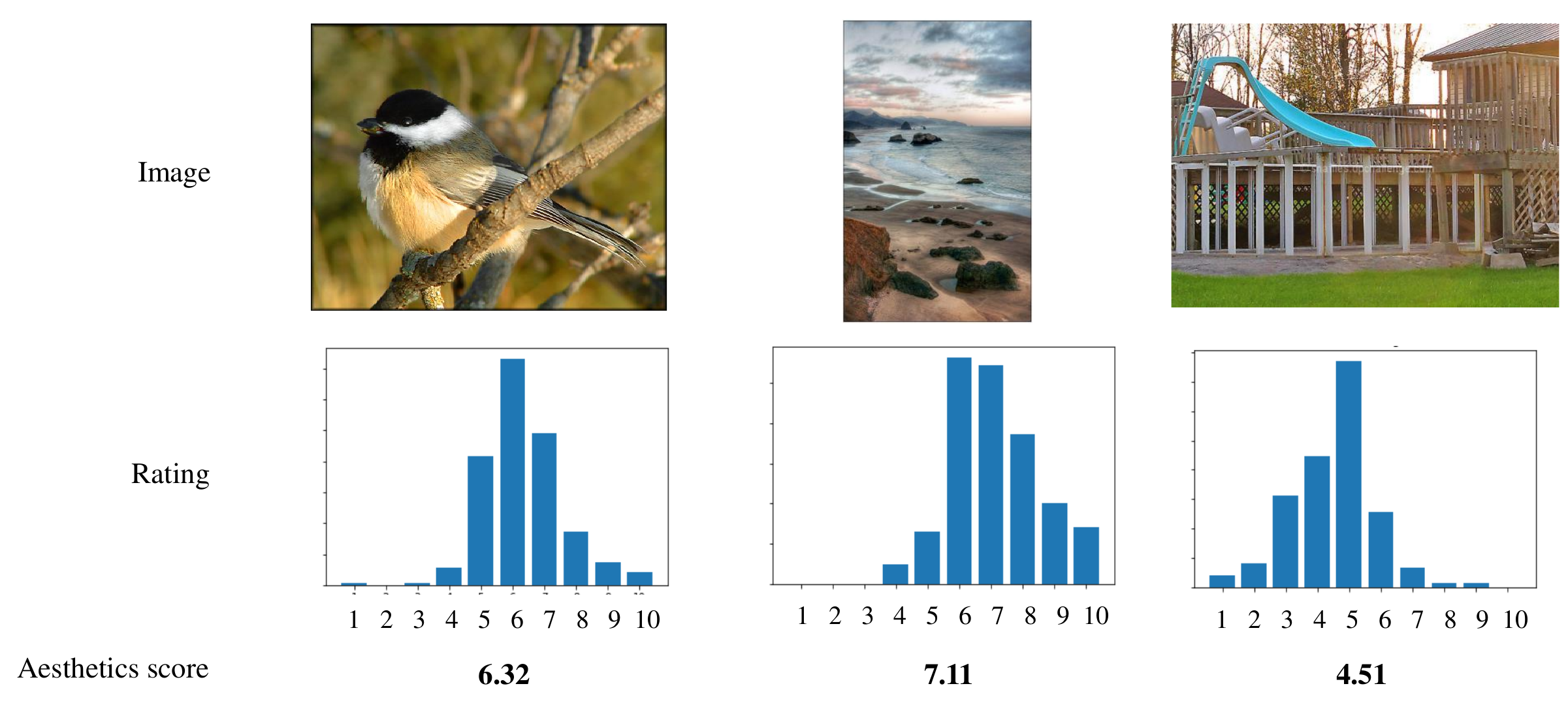}
  }
  \caption{Sample images (top), normalized rating histograms (middle), and means of the rating histograms calculated as aesthetics scores (bottom) taken from the AVA dataset.}
  \label{fig:ava_sample}
\end{figure*}

\section{Introduction}

\IEEEPARstart{T}{he}
spread of social networking services (SNS) has increased the importance of posting attractive images to make messages more influential. This applies to both individual and business uses of SNS. However, because most users do not have the required skills to select, edit, and generate aesthetic images, there is currently a strong need for an automatic process for obtaining aesthetic images. To realize such a process, it is essential to be able to automatically assess image aesthetics precisely.

In general, for image aesthetic assessment, it is important to effectively extract features from the entire image and combine them, because aesthetics stems from human subjectivity, which makes aesthetics assessment different from other recognition tasks. In the early approaches~\cite{ke2006design, datta2006studying, luo2008photo, marchesotti2011assessing, li2010towards, dhar2011high, luo2011content, Lo2012assessment, karayev2014recognizing}, handcrafted features were adopted, including object composition, space complexity, and color harmony. Following these studies, based on the success of convolutional neural networks (CNNs) on object recognition tasks, many researchers~\cite{lu2014rapid, lu2015rating, fu2018image, kao2017deep, chopra2005learning, shen2017deep, schroff2015facenet, kong2016photo, ko2018pac, schwarz2018will, zhou2016joint, mai2016composition, kao2015visual, jin2016image, talebi2018nima, wang2019mpemd, zhang2019gated, lee2019image} adopted CNNs as feature extractors.

Aside from the features contained images themselves, additional information is generally also included to improve prediction accuracy, such as scene or style annotations in datasets~\cite{lu2014rapid, lu2015rating, fu2018image, kao2017deep, kong2016photo, mai2016composition}, multimodal text comments~\cite{zhou2016joint}, object tags~\cite{roy2018predicting}, and saliency maps~\cite{ma2017lamp}.
Although these additional characteristics improve the performance of aesthetic assessment methods, they result in high cost when creating new datasets and limitation on application to other tasks because specific additional information is required during the training phase or sometimes the evaluation phase. In this study, we focus on a fundamental and versatile approach for effective image feature extraction to perform aesthetics assessment. Therefore, we only use images to predict aesthetics scores during both training and evaluation.

Previous research has focused on three kinds of image aesthetics assessment tasks: positive/negative binary classification tasks~\cite{luo2011content, Lo2012assessment, lu2014rapid, lu2015rating, kao2017deep, kong2016photo, schwarz2018will, zhou2016joint, mai2016composition, lu2015deep, ma2017lamp, sheng2018attention}, aesthetics rating distribution prediction tasks~\cite{cui2017distribution, cui2018distribution, wang2017image, fang2018image, jin2018predicting}, and aesthetics score prediction tasks~\cite{roy2018predicting, kao2015visual, jin2016image, talebi2018nima, wang2019mpemd, zhang2019gated, lee2019image}. Positive/negative classification tasks have been tackled in the most studies, and the numbers of studies for aesthetics rating distribution prediction and aesthetics score prediction tasks are relatively small. 
In this paper, we focus on aesthetics \textit{score prediction}, which is a task to predict the mean of the aesthetics rating distribution of an image. The aesthetics rating distribution is generated from human votes. Sample images, normalized rating distributions, and the calculated aesthetic scores of a large-scale aesthetics dataset, called the AVA dataset~\cite{murray2012ava}, are shown in \Fref{fig:ava_sample}. 
Aesthetics score prediction is useful for applications that require quantitative evaluations, such as image recommendation~\cite{niu2018neural} and photo selection~\cite{shen2018photo} for advertisements~\cite{xia2019deep}. 
The aesthetics score prediction expands the way of practical applications compared to the aesthetics positive/negative binary classification.
Furthermore, 
the aesthetics scores predicted by models can be applied as references
for image processing tasks, including image cropping~\cite{wang2017deep}, image retargeting~\cite{avidan2007seam}, and image color enhancement~\cite{wang2011example}.

Studies on aesthetics score prediction have been conducted by Kao \etal~\cite{kao2015visual}, Jin \etal~\cite{jin2016image}, Roy \etal~\cite{roy2018predicting}, Talebi \etal~\cite{talebi2018nima}, Zhang \etal~\cite{zhang2019gated}, and Lee \etal~\cite{lee2019image}. However, in all existing methods, it is necessary to rescale images into square (or at least fixed-size) images regardless of their original image aspect ratios. This is true even for the most outstanding method of these method, called neural image assessment (NIMA), which was proposed by Talebi \etal~\cite{talebi2018nima}. The lack of aspect ratio information for the original images and contents of them can affect the prediction of aesthetics scores, especially for images having unusual aspect ratios. Furthermore, it can easily cause contradictions with human aspect-ratio-dependent aesthetics.

To address this issue, we propose an aspect-ratio-preserving patch-learning approach for aesthetics score prediction. This approach consists of cropping several patches from an input image, predicting normalized aesthetics rating distributions for each patch, and calculating the final aesthetics score by using these distributions. In the training phase, we use the \textit{collective} / \textit{individual} earth mover's distance (EMD) as a part of the loss function. Experimental results obtained using the AVA dataset~\cite{murray2012ava}, which has more than 250,000 images, demonstrate that the proposed aspect-ratio-preserving patch-learning method outperforms other aesthetics score prediction methods. Our method, named MPA-Net (which stands for multi-patch aggregation network), yields a linear correlation coefficient (LCC) of aesthetics scores 0.073 (11.5\%) higher and a Spearman's rank correlation coefficient (SRCC) 0.088 (14.4\%) higher than those of NIMA,~\cite{talebi2018nima}, which was used as a baseline method.
Furthermore, compared with GPF-CNN~\cite{zhang2019gated}, which is the state-of-the-art aesthetics score prediction method, MPA-Net achieved slightly better results in terms of LCC and SRCC and also yielded a 0.0115 (4.18\%) lower mean squared error (MSE). 
Another advantage of our method is that the MSEs for predictions made for extraordinarily vertically or horizontally long images are significantly lower compared with the baseline.

In summary, the main contributions of this study are as follows:
\begin{itemize}
    \item We propose an aspect-ratio-preserving patch-learning approach for predicting aesthetics scores that preserved the original aspect ratio of contents to make more accurate predictions.
    \item Experimental results demonstrate that the proposed model, MPA-Net, achieves an LCC of aesthetics scores and an SRCC that are 0.073 (11.5\%) and 0.088 (14.4\%) higher, respectively, compared with the NIMA baseline~\cite{talebi2018nima}. Moreover, the obtained MSE is at least 0.0115 (4.18\%) lower compared with that of existing methods. In particular, our method performs significantly better than other approaches for images with unusual aspect ratios.
    \item Our widely applicable method uses images and aesthetic ratings without requiring additional information to achieve high performance when predicting aesthetic scores. This makes it applicable to other datasets and other tasks which requires maintaining original aspect ratios of contents.
\end{itemize}


\section{Related Works}

\subsection{Aesthetics Assessment}
Aesthetics assessment can be broadly categorized into three types of tasks: positive/negative aesthetic binary classification, aesthetics rating distribution prediction, and prediction of the mean of the rating distribution. 

\vspace{2mm}\noindent\textbf{Aesthetic Binary Classification. \ \ } 
Positive/negative binary classification has long been studied. Initially, it was tackled by Ke \etal~\cite{ke2006design} and Datta \etal~\cite{datta2006studying} by heuristic features, such as colors, shapes, and textures. Following them, several studies~\cite{luo2008photo, li2010towards, marchesotti2011assessing, dhar2011high, luo2011content, Lo2012assessment, karayev2014recognizing} have challenged this task using elaborately designed features and improved machine learning methods.
In such a situation, Lu \etal~\cite{lu2014rapid} first adopted deep neural networks for image aesthetics assessment including binary aesthetics classification. After that study, deep neural networks have been widely adopted in image aesthetics assessment. Lu \etal~\cite{lu2014rapid} also used a global view and a local view from a photo as the input of the model. This global/local input approach has been developed to multi-patch approaches~\cite{lu2015deep, ma2017lamp, sheng2018attention}. Also, pairwise learning~\cite{kong2016photo, schwarz2018will, lee2019image} and training with additional information, such as photo categories and comments~\cite{zhou2016joint, mai2016composition, kao2017deep}, has been examined recently.

\noindent\textbf{Aesthetics Rating Distribution Prediction. \ \ } 
On the other hand, the aesthetics rating distribution prediction task just has a short history. For this task, several studies~\cite{cui2017distribution, cui2018distribution, wang2017image, fang2018image, jin2018predicting} have been conducted. They generally adopted CNNs as feature extractors and designed their loss functions to improve the performance. For instance, hinge loss was adopted by Cui \etal~\cite{cui2017distribution}, Kullback-Leibler (KL) divergence was adopted by Cui \etal~\cite{cui2018distribution}, Wang \etal~\cite{wang2017image}, and Fang \etal~\cite{fang2018image}, and Jensen-Shannon divergence was adopted by Jin \etal~\cite{jin2018predicting}.

\noindent\textbf{Aesthetics Score Prediction. \ \ } 
The prediction of the mean of the rating distribution has been conducted more widely than the prediction of aesthetics rating distribution, but not as popular as aesthetics binary classification. The mean of the rating distribution is usually called ``aesthetics score.'' From here, we will explain previous works related to the task we focus on: aesthetics score prediction.

\vspace{\baselineskip}

To the best of our knowledge, the first attempt at predicting aesthetics scores was made by Kao \etal~\cite{kao2015visual} using a regression network. This network comprised five convolution layers and four fully connected (FC) layers and directly predicted the aesthetics scores of images. Jin \etal~\cite{jin2016image} trained a network by adding large weights to images with rare aspect ratios in the dataset. Roy \etal~\cite{roy2018predicting} also employed additional object tags to predict aesthetics scores. In contrast with these methods, instead of directly calculating aesthetic score via regression, Talebi \etal~\cite{talebi2018nima} proposed NIMA, an approach that calculates aesthetics scores from predicted aesthetics rating distributions. NIMA has two outstanding novelties. The first is that NIMA employs rating distributions to utilize more information about ratings compared with direct aesthetics score regression. The second is that NIMA adopted the earth mover's distance (EMD)~\cite{levina2001earth, hou2016squared} for training its parameters. EMD is a distribution distance function that considers inter-class relationships. Therefore, NIMA can learn the global characteristics of distributions without sticking to elaborately fitting the local values of distributions.

However, owing to the restriction of CNNs, all images have to be rescaled to square images to be fed into the network regardless of their aspect ratios. Through this transformation, images lose their original aspect ratio information of contents, which can affect the prediction of aesthetics scores, especially for images with unusual aspect ratios. As a result, this creates a contradiction in the fact that the NIMA network predicts the same aesthetics score for both the original and the rescaled images, whereas humans can easily detect a decrease in aesthetics for the rescaled images.

Zhang \etal~\cite{zhang2019gated} also proposed the method to utilize saliency maps. Lee \etal~\cite{lee2019image} adopted a pairwise comparison model for aesthetics score prediction with discrete values, while the predictions by the other methods were continuous. Though the method proposed by Lee \etal~\cite{lee2019image}  performed outstanding results, the prediction with discrete values has the restriction of applications such as consecutive image editing to increase aesthetics. Therefore, we classify this method in a different category and not directly compare it with our method.

\subsection{Aspect-ratio-preserving Aesthetics Assessment}
To solve the above-mentioned problem, several studies for performing aesthetics assessment while preserving the original aspect ratios of contents have been conducted.
Mai \etal~\cite{mai2016composition} and Cui \etal~\cite{cui2018distribution} employed adaptive spatial pooling and global average pooling, respectively. Lu \etal~\cite{lu2014rapid, lu2015deep}, Ma \etal~\cite{ma2017lamp}, and Sheng \etal~\cite{sheng2018attention} used a multi-patch approach. Zhang \etal~\cite{zhang2019gated} also adopted a multi-patch approach, but they did not focus on preserving aspect ratios of contents.
Lu \etal~\cite{lu2015deep} demonstrated that spatial pyramid pooling (SPP), which is a kind of pooling strategy, did not bring significant contributions for aesthetics binary classification.
Furthermore, by using multi-patch approaches, it is easier to perform batch training, which improves training time and trained model performance, than when using pooling strategies.
Thus, we also adopted multi-patch training and evaluation. 

Among these multi-patch methods, Sheng \etal~\cite{sheng2018attention} proposed a weighted aggregation system for multiple patches with the original aspect ratio of contents, which is the most recent highly effective method. Using this system, the network can be trained strongly from wrongly predicted patches. However, multi-patch learning has only been applied for aesthetic binary classification. We employed aspect-ratio-preserving multi-patch learning to predict aesthetics scores by predicting normalized aesthetics rating distributions. 
A brief comparison of the functions of NIMA~\cite{talebi2018nima}, $\MP{ada}$ (proposed by Sheng \etal), and the proposed method is shown in \Tref{tab:prev_works_comp}.

This study extends our previous study~\cite{wang2019mpemd}. In the previous study, only \textit{random} multi-patch tests were conducted. In this paper, we present the results of \textit{fixed location} multi-patch tests and detailed ablation studies on the components used in the proposed model~\cite{wang2019mpemd}.

\begin{table}[!tb]
  \caption{Comparison of functions among previous aesthetics assessment works and our method.
  }
  \begin{center}
  \begin{tabular}{c|ccc}
\hline
 & NIMA~\cite{talebi2018nima} & $\MP{ada}$~\cite{sheng2018attention} & \it{Proposed} \\
\hline
Score prediction & $\checkmark$ & & $\checkmark$ \\
Preservation of aspect ratio &  & $\checkmark$ & $\checkmark$ \\
\hline
\end{tabular}
\end{center}
\label{tab:prev_works_comp}
\vspace{-10pt}
\end{table}


\section{Multi-Patch Aggregation Network (MPA-Net)}
In this section, we introduce our training and prediction system for assessing aesthetics scores. We describe training and test phases, and the proposed loss functions in detail.

\begin{figure*}[bt]
  \centering
  \scalebox{0.46}{
  \includegraphics{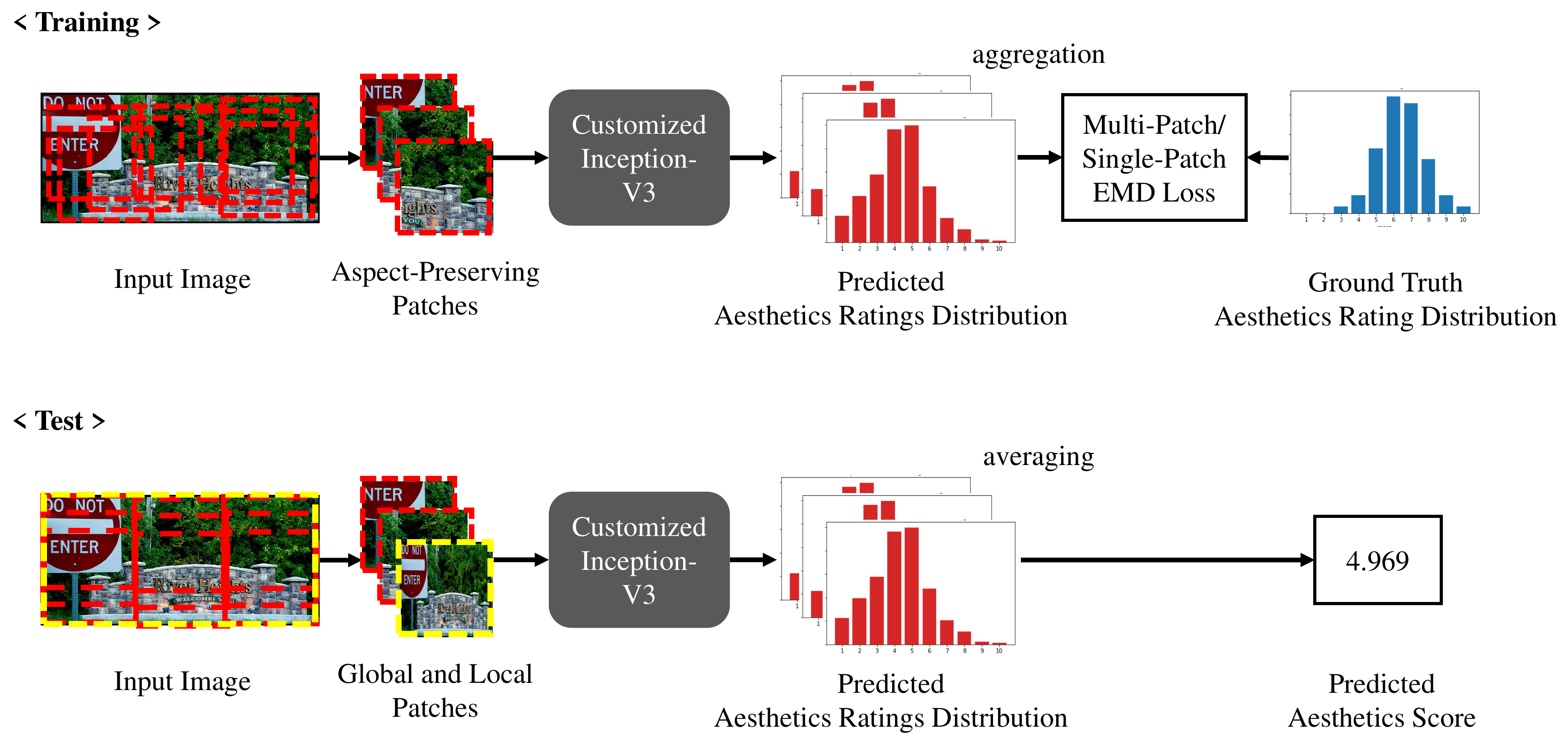}
  }
  \caption{Aspect-ratio-preserving patch training/evaluation flowcharts of our method. The yellow patch in the test stage indicates the global patch.}
  \label{fig:mpemd_network}
\end{figure*}

\subsection{Aspect-ratio-preserving Patch Training} 
In the training phase, square patches are cropped at random from an input image without altering the aspect ratio of its contents. By extracting patches with the original aspect ratio of contents, the model can learn to perform image feature extraction with the same aspect ratio of contents that humans see. Therefore, it is considered to be easier for the model to learn about the human subjectivity of aesthetics. Furthermore, the model is expected to be trained effectively because no disturbances are caused by uniform square reshaping processes that ignore the original aspect ratio of contents, which happened in related methods, such as NIMA~\cite{talebi2018nima}. The extracted aspect-ratio-preserving patches are fed into the model and the distributions of the aesthetics ratings are predicted for each patch. The sum of each distribution is normalized to 1 by calculating a softmax function over the output of the last FC layer. The EMD (\Eref{eq:emd}, described in detail in~\Sref{sec:emd}) is calculated for each rating distribution. During training, the loss value of each image is computed using EMD through the loss functions described in \Sref{sec:loss_func}. The model parameters are updated via backpropagation from these loss values; updates are repeated for several epochs using different cropped patches. A flowchart of the MPA-Net method is shown in \Fref{fig:mpemd_network}.

\subsection{Loss Function}\label{sec:loss_func} 
\noindent\textbf{Earth mover's distance (EMD).\ \ }\label{sec:emd} 
For the distance function between the rating distributions, we employ earth mover's distance (EMD), similarly to NIMA~\cite{talebi2018nima}. EMD is a distance function for determining the distance between two distributions. Unlike cosine similarity or KL divergence, EMD can consider distance among classes. Therefore, the model can learn the global properties of the rating distributions, without being limited to elaborately fitting the local values of each class. The $r$-norm EMD distance is defined as the minimum cost of transporting values from one distribution to the another, where the distance between the $i$-th class $s_i$ and the $j$-th class $s_j$ is calculated as ${\| s_i - s_j \|}_r$, under the assumption that the two distributions have the same classes in the same order.

For $N$-class aesthetics ratings, if the value of the $i$-th rating class $s_i$ is $i$, where $1 \leq i \leq N$, the distance between the $i$-th rating class $s_i$ and the $j$-th class $s_j$ is calculated as ${|i-j|}^r$. In this case, as demonstrated by Levina \etal~\cite{levina2001earth}, the $r$-norm EMD between two normalized aesthetics rating distributions is calculated as follows:
\begin{eqnarray}
{\rm EMD}^{(r)} = \left( \frac{1}{N}\sum_{k=1}^{N} \left| {\rm CDF_{\bf p}} (k)-{\rm CDF_{\bf \hat{p}}} (k) \right| ^ r \right) ^ {\frac{1}{r}}, \label{eq:emd}
\end{eqnarray}
where ${\rm CDF_{{\bf p} / {\bf \hat{p}}}} (k)$ denotes the cumulative distribution function of the ground-truth rating distribution ${\rm\bf p}$ and the predicted rating distribution ${\rm\bf \hat{p}}$, which are defined as $\sum_{k=1}^{N} {\rm\bf p}$ and $\sum_{k=1}^{N} {\rm{\bf \hat{p}}}$, respectively. We set $r$ to 2, as done in NIMA.

\vspace{2mm}\noindent\textbf{Training patch aggregation. \ \ } \label{sec:patch_aggregation}
We refer to the method proposed by Sheng \etal~\cite{sheng2018attention} for multi-patch weighted aggregation, which outperforms the other previous works in positive/negative aesthetic binary classification tasks. Compared with the loss function used by Sheng \etal, we adopt logarithmic 2-norm EMD (${\rm EMD^{(2)}}$, hereinafter referred to as EMD) to calculate the loss of the predicted rating distributions in place of the log probability for binary classification. We employed logarithmic EMD instead of simply EMD because we expected a logarithmic function would accelerate training.

We designed two training strategies and their corresponding loss functions.
The first strategy is to crop a \textit{collection} of patches randomly from each image in every epoch and minimize the loss value calculated by aggregating the EMDs from these patches.
The first \textit{collective} strategy uses the same multi-patch approach as the method proposed by Sheng \etalp.
The second strategy is to crop one patch at random from each image in every epoch and directly minimize the loss value from that patch. With this strategy, a model is trained by multiple patches \textit{individually}. To train the model with many patches, the number of epochs for this \textit{individual} strategy was set larger than that for the \textit{collective} strategy. 

Both strategies adopt the logarithm of the EMD and weight coefficients.
We named the loss function of the \textit{collective} strategy ``$\COLEMD{}$'' and the loss function of the \textit{individual} strategy ``$\INDEMD{}$.'' 
They are defined as follows:
\begin{eqnarray} 
    {\rm \COLEMD{}}(\mathcal{P}) & = & - \frac{1}{|\mathcal{P}|} \sum_{p \in \mathcal{P}} \omega_\beta \cdot \log \left( {\rm EMD_c}(p) \right), \label{eq:colemd} \\
    {\rm \INDEMD{}}(p) & = & - \omega_\beta \cdot \log \left( {\rm EMD_c}(p) \right), \label{eq:indemd} 
\end{eqnarray}
where 
$\mathcal{P}$ is a set of cropped square patches from the original image, $p$ denotes a single cropped patch, and $\EMDc$ is a variable converted from the original EMD to represent a kind of certainty of predicted rating distributions.
The purpose of training is to minimize the EMD, which is equivalent to maximizing $\EMDc$. $\EMDc$ is defined as follows: 
\begin{eqnarray}
    {\rm EMD_c}(p)  \!=\!  \left\{
    \begin{array}{ll}
        \epsilon, & (1 - k \cdot {\rm EMD} < \epsilon) \\
        1 \!-\! k \!\cdot\! {\rm EMD}, & (\epsilon \!\leq\! 1 - k \!\cdot\! {\rm EMD}) 
    \end{array} \right. \label{eq:emd_c} 
\end{eqnarray}
where $\epsilon$ is an appropriately small positive constant and $k$ is an expansion coefficient. $\EMDc$ takes values close to 1 when EMD is small and near 0 when EMD is large. The value of $\EMDc$ is restricted to $\left[\epsilon,\ 1\right]$. The hyperparameter $k$ is used to adjust the sensitivity of the converted certainty variable $\EMDc$ to EMD. 
As $k$ increases, the variation of EMD causes a larger change in $\EMDc$.

\begin{figure}[t]
  \centering
  \scalebox{0.5}{
  \includegraphics{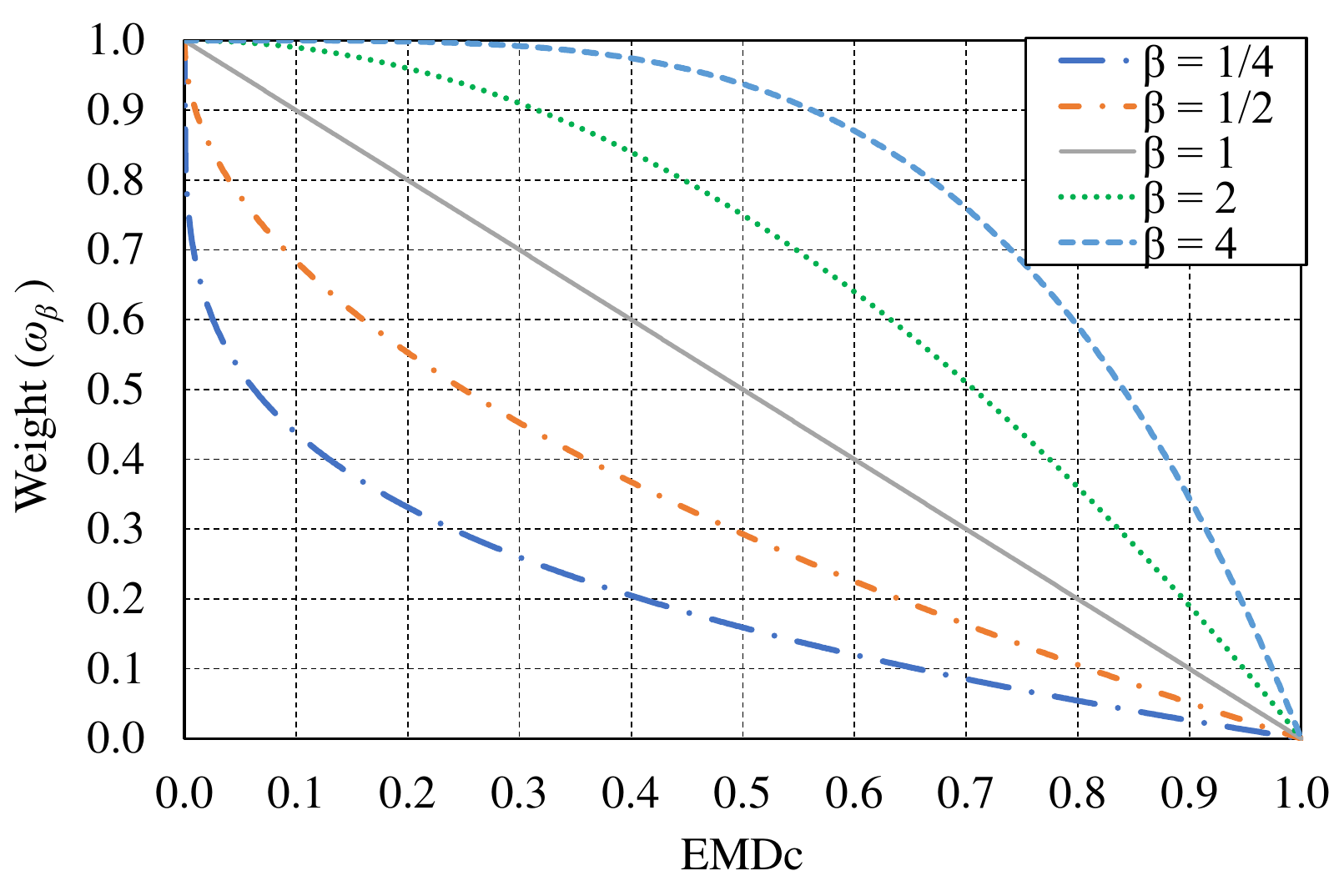}
  }
  \caption{Relationship between the patch weight $\omega_\beta$ and the certainty variable $\EMDc$, with respect to various values of $\beta$.
  }
  \label{fig:mp_ada_wb}
\end{figure}

$\omega_\beta$ is introduced as the weight of the patches and is defined as
\begin{eqnarray}
    \omega_\beta = 1 - {\rm EMD_c}^\beta. \label{eq:omega_beta}
\end{eqnarray}
$\omega_\beta$ is high when the certainty variable $\EMDc$ is low, and vice versa. This means that $\omega_\beta$ is large when EMD is large. The value of $\omega_\beta$ ranges from 0 to 1. The hyperparameter $\beta\ (\beta>0)$ determines the range of $\EMDc$ with which the patches are trained strongly.
\Fref{fig:mp_ada_wb} shows how the patch weight $\omega_\beta$ varies with the certainty variable $\EMDc$ for various values of $\beta$. For example, as shown in \Fref{fig:mp_ada_wb}, if $\beta$ is large, even a patch with a large $\EMDc$ will be weighted heavily. This means that a patch with small EMD will also be strongly trained when $\beta$ is large.

The effects of $k$ and $\beta$ are dependent on each other; thus, $k$ and $\beta$ should be optimized jointly.

\subsection{Test Patch Aggregation Flow} \label{sec:test_flow}
Multi-patch evaluation is conducted in the prediction stage. Unlike in the training phase, patches at \textit{fixed} locations are cropped from the input image. First, $m \times m$ local patches are cropped at equal intervals. The entire image is also resized to the square size and used as the global patch. Though global patch does not maintain the original aspect ratios of images, it serves supplementally to reflect the overall view of the image. An example of the cropping process for test is shown in \Fref{fig:mpemd_network}. Then, the predicted rating distribution of the input image is calculated as the simple average of the normalized rating distributions predicted from the cropped and resized patches. The aesthetics score is computed as the mean of the averaged rating distributions.

\begin{figure}[bt]
  \centering
  \scalebox{0.55}{
  \includegraphics{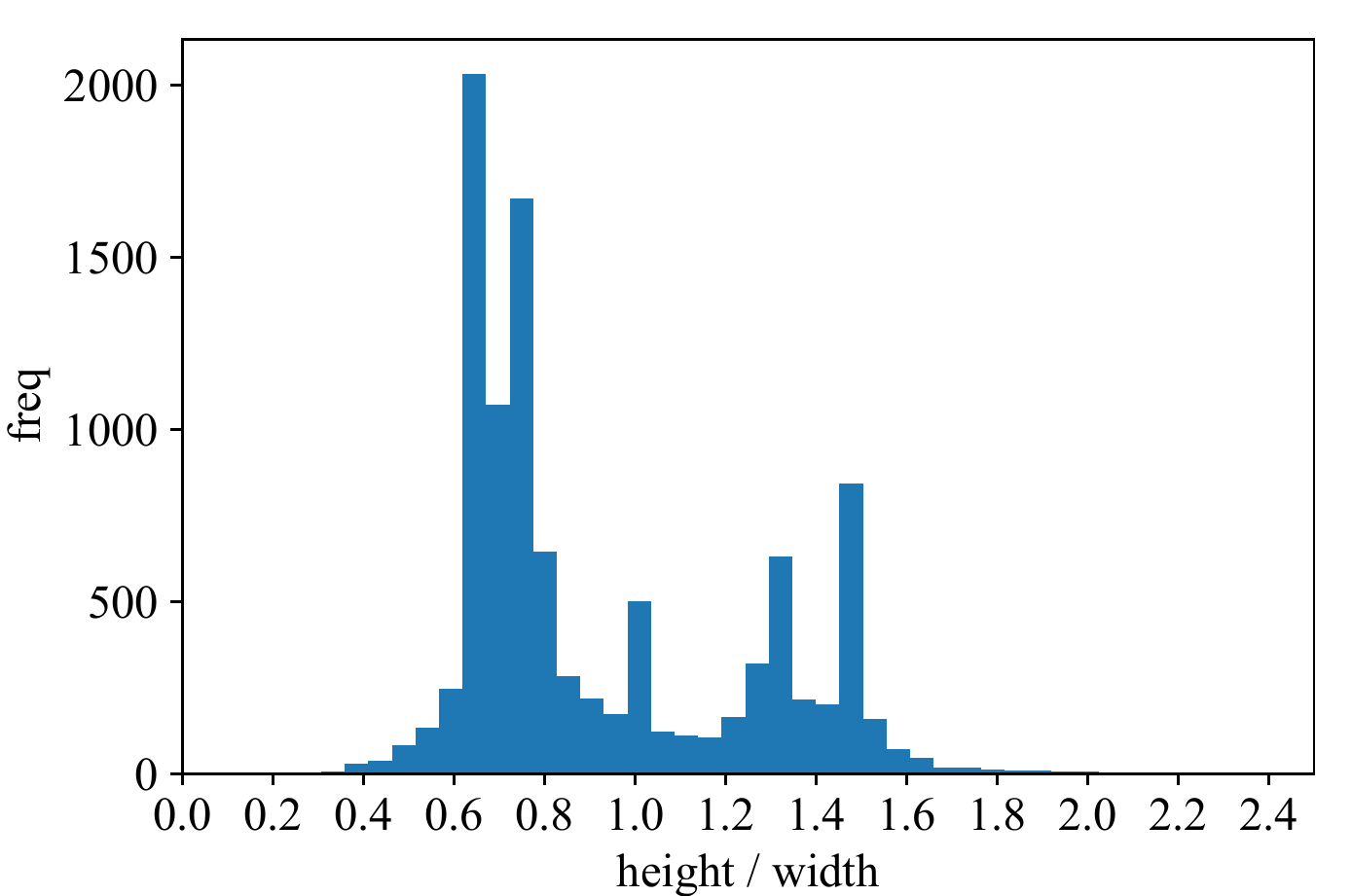}
  }
  \caption{Histogram of the aspect ratios (height/width) of images sampled from the AVA dataset.}
  \label{fig:ava_aspect_ratio_histgram}
\end{figure}

\begingroup
\renewcommand{\arraystretch}{1.5}
\begin{table*}[!tb]
  \caption{Definitions of loss functions analyzed in ablation studies.
  }
  \begin{center}
  \begin{tabular}{l|ll|l}
\hline\hline
\tabcenter{Loss function} & \tabcenter{(i) Log $\EMDc$ } & \tabcenter{(ii) Weight coef. $\omega_\beta$ } & \tabcenter{Definition}   \\
\hline
$\COLEMD{simple}$    &               &               & $- \frac{1}{|\mathcal{P}|} \sum_{p \in \mathcal{P}}  {\rm EMD_c}(p)$ \\
$\COLEMD{weighted}$  &               & $\checkmark$  & $- \frac{1}{|\mathcal{P}|} \sum_{p \in \mathcal{P}} \omega_\beta \cdot {\rm EMD_c}(p) $ \\
$\COLEMD{log}$       & $\checkmark$  &               & $- \frac{1}{|\mathcal{P}|} \sum_{p \in \mathcal{P}}  \log \left( {\rm EMD_c}(p) \right)$ \\
$\COLEMD{}$          & $\checkmark$  & $\checkmark$  & $- \frac{1}{|\mathcal{P}|} \sum_{p \in \mathcal{P}} \omega_\beta \cdot \log \left( {\rm EMD_c}(p) \right)$ \\
\hline 
$\INDEMD{simple}$    &               &               & $- {\rm EMD_c}(p) $ \\
$\INDEMD{weighted}$  &               & $\checkmark$  & $- \omega_\beta \cdot {\rm EMD_c}(p)$ \\
$\INDEMD{log}$       & $\checkmark$  &               & $- \log \left( {\rm EMD_c}(p) \right)$ \\
$\INDEMD{}$          & $\checkmark$  & $\checkmark$  & $- \omega_\beta \cdot \log \left( {\rm EMD_c}(p) \right)$ \\
\hline\hline
\end{tabular}
\end{center}
\label{tab:ablation_loss_functions}
\vspace{-10pt}
\end{table*}
\endgroup


\section{Experiment}
In this section, we first describe the dataset used in our experiment. 
Then, we introduce the model architecture as well as the pre-training conducted only for the \textit{collective} training strategy described in \Sref{sec:patch_aggregation}.
Finally, we explain the setup for the ablation study conducted to verify both the loss functions and the test flow.

\subsection{Dataset} \label{sec:dataset}
We trained and evaluated the MPA-Net using the AVA dataset~\cite{murray2012ava}. The AVA dataset comprises 250,000 images collected from the online photography community website \url{www.dpchallenge.com}. Each image is associated with 10-stages ratings, ranging from 1 to 10. The number of raters assigned to each image ranges from 78 to 649, and the average value is 210. Samples of the AVA dataset, including images, normalized rating histograms, and the means of the rating histograms (called as aesthetic scores) are shown in \Fref{fig:ava_sample}. Aside from the ratings, some images have additional attributes, such as semantic and photographic style information, which were not used neither for training nor testing in our experiment. 

\Fref{fig:ava_aspect_ratio_histgram} shows the histogram of the aspect ratios (height/width) of the images in the AVA dataset. As shown in \Fref{fig:ava_aspect_ratio_histgram}, most images have aspect ratios rating from 0.6 to 0.8. In particular, there are two peaks within the ranges of 0.62 to 0.67 and 0.72 to 0.77. This concentration can be explained by the fact that normal digital cameras are configured to take photos with height/wdith ratios of 2:3 (the aspect ratio is 0.66) or 3:4 (the aspect ratio is 0.75). In addition, several frequency peaks can be observed for height/width ratios of 1:1 (the aspect ratio is 1.0), 4:3 (the aspect ratio is 1.33), and 3:2 (the aspect ratio is 1.5). 
In other words, the AVA dataset contains a relatively small number of images with aspect ratios not falling within the range described above, which means that those uncommon aspect ratios have less training images.

We used the AVA dataset~\cite{murray2012ava} for both training and evaluation. The AVA dataset we used contains 255,494 pairs of an image and a rating histogram. In the same way as the previous multi-patch methods~\cite{mai2016composition, lu2015deep, sheng2018attention}, we used 92 \% of the entire dataset for training. Additionally, half of the remaining dataset (4\% of the entire dataset) was used for testing and the other half was used for validation. Therefore, 235,054 images were used for training, 10,220 images were used for validation, and the remaining 10,220 images were used for testing. 
It should be noted that some methods from previous works used different numbers of images for the training/validation/test datasets. For example, Kao \etal~\cite{kao2015visual}, Jin \etal~\cite{jin2016image}, and Roy \etal~\cite{roy2018predicting} used approximately 250,000 images for training and 5,000 images for testing, whereas Talebi \etal~\cite{talebi2018nima} used approximately 204,000 images for the training of NIMA and 51,000 images for testing it. The reason we chose the above-described proportion (92:4:4) is that 5,000 test images were not enough for our analysis on aspect ratios described in \Sref{sec:results}, and 51,000 images were too many for testing. To make a fair comparison, we also show the results of the reimplemented NIMA trained method with 92\% of the entire AVA dataset in \Sref{sec:results}.

\subsection{Model Architecture and Pre-training}

We used the same model architecture as that used in NIMA~\cite{talebi2018nima}, namely a customized Inception-V3~\cite{szegedy2016rethinking} network with the last FC layer replaced by a randomly initialized FC layer with 10 output channels. All layers except for the last new FC layer were initialized with the parameters pre-trained on the ImageNet dataset~\cite{russakovsky2015imagenet}.

Before training using the proposed MPA-Net \textit{collective} training strategy, we pre-trained the model with square rescaled images converted from the AVA~\cite{murray2012ava} training set as done for NIMA. This pre-training was conducted to rapidly learn global features from entire images. We adopted this process only for \textit{collective} training because it tends to take longer to converge compared with \textit{individual} training. 
In the pre-training stage, all images from the training set were resized to $342\times342$, after which $299\times299$ random cropping and random horizontal flipping were performed for data augmentation. We set the learning rate to ${10}^{-3}$ instead of $3\times{10}^{-7}$ or $3\times{10}^{-6}$, which were the values reported by Talebi \etal~\cite{talebi2018nima}, because the model could not be trained adequately in our setup using those learning rates. Additionally, we used a momentum SGD optimizer with a momentum of 0.9 and let learning rate decay by a factor of 0.95 after every 10 epochs. We trained the model for 100 epochs.

\begin{table}[!tb]
  \caption{Training parameters for loss functions shown in \Tref{tab:ablation_loss_functions}.
  }
  \begin{center}
  \begin{tabular}{l|lll|l}
\hline\hline
\tabcenter{Loss function} & \multicolumn{3}{|c|}{Learning rate} & \tabcenter{Epoch} \\
\cline{2-4} 
& Init. rate & Decay factor & Decay intvl. & \\
\hline
$\COLEMD{simple}$    & ${10}^{-4}$ & 0.85 & 5 [epoch] & 50 \\
$\COLEMD{weighted}$  & ${10}^{-3}$ & 0.85 & 5 [epoch] & 50 \\
$\COLEMD{log}$       & ${10}^{-3}$ & 0.7 & 10 [epoch] & 50 \\
$\COLEMD{}$          & ${10}^{-3}$ & 0.7 & 10 [epoch] & 50 \\
\hline 
$\INDEMD{simple}$    & ${10}^{-3}$ & 0.9 & 10 [epoch] & 200 \\
$\INDEMD{weighted}$  & ${10}^{-2}$ & 0.9 & 10 [epoch] & 200 \\
$\INDEMD{log}$       & ${10}^{-3}$ & 0.9 & 10 [epoch] & 200 \\
$\INDEMD{}$          & ${10}^{-2}$ & 0.9 & 10 [epoch] & 200 \\
\hline\hline
\end{tabular}
\end{center}
\label{tab:train_hyperparameter}
\vspace{-10pt}
\end{table}

\begin{table*}[!tb]
  \caption{Comparison of the aesthetics score prediction performance of our methods, including ablation experiments. The rows correspond to different training loss functions and the columns indicate the test patch-selection strategy used. The best values are shown in bold for each metric, for the $\COLEMD{}$ and $\INDEMD{}$ series.
  }
  \begin{center}
  \begin{tabular}{l|ll|ll|ll}
\hline\hline
\tabcenter{Experiment} & \multicolumn{2}{|c|}{MP-Random} & \multicolumn{2}{|c|}{MP-Local} & \multicolumn{2}{|c}{MP-GlobalLocal} \\
\cline{2-7} 
& LCC $\uparrow$ & RMSE $\downarrow$ & LCC $\uparrow$ & RMSE $\downarrow$ & LCC $\uparrow$ & RMSE $\downarrow$ \\
\hline
$\COLEMD{simple}$    & 0.6815    & 0.5357    & 0.6918    & 0.5274    & 0.6954    & 0.5249 \\
$\COLEMD{weighted}$  & 0.6935    & 0.5330    & 0.7019    & 0.5245    & \textbf{0.7043}    & \textbf{0.5167} \\
$\COLEMD{log}$       & 0.6900    & 0.5280    & 0.6986    & 0.5210    & 0.7012    & 0.5189 \\
$\COLEMD{}$          & 0.6923    & 0.5257    & 0.7009    & 0.5190    & 0.7038    & 0.5172 \\
\hline 
$\INDEMD{simple}$    & 0.6960    & 0.5225    & 0.7045    & 0.5163    & 0.7074    & 0.5150 \\
$\INDEMD{weighted}$  & 0.6975    & 0.5216    & 0.7062    & 0.5151    & 0.7089    & 0.5138 \\
$\INDEMD{log}$       & 0.6966    & 0.5222    & 0.7047    & 0.5160    & 0.7072    & 0.5144 \\
$\INDEMD{}$          & 0.6985    & 0.5212    & 0.7068    & 0.5149    & \textbf{0.7096}    & \textbf{0.5135} \\
\hline\hline
\end{tabular}

\end{center}
\label{tab:result_ablation_studies}
\vspace{-10pt}
\end{table*}

\subsection{Experiment Configuration} \label{sec:ablation_training}

\vspace{2mm}\noindent\textbf{Training. \ \ }
For training, we randomly cropped patches in the following manner. First, we rescaled the shorter edge of every image in the dataset to 342 pixels while maintaining its aspect ratio. Then, we extracted $299\times299$ croppings from each rescaled image. When training using $\COLEMD{}$, we cropped eight patches from each image at the same time. On the other hand, when using $\INDEMD{}$, the training process required only one patch at one epoch. The patch/patches were cropped and used only once in each epoch, and different patch/patches were prepared for different epochs.

To investigate the effectiveness of each component of the loss function, in addition to analyzing the proposed loss functions $\COLEMD{}$ (\Eref{eq:colemd}) and $\INDEMD{}$ (\Eref{eq:indemd}), we conducted ablation studies to examine the effect of (i) the logarithm of $\EMDc$ and (ii) the weight coefficient $\omega_\beta$ for $\EMDc$. The definitions of these loss functions are shown in \Tref{tab:ablation_loss_functions}.

For the hyperparameters of the loss function, we set the $k$ used in $\EMDc$ to 1.2 and the $\beta$ used in $\omega_\beta$ to 0.4, based on hyperparameter tuning using the tree-structured Parzen estimator (TPE)~\cite{bergstra2011algorithms} implemented by Optuna~\cite{optuna}. For the optimizer, we used a momentum SGD optimizer with a momentum value of 0.9 and a weight decay rate of ${10}^{-4}$. The other training parameters are shown in \Tref{tab:train_hyperparameter}: initial learning rate, learning rate decay factor, learning rate decay interval, and learning epochs. All models were implemented using PyTorch v.0.4.0~\cite{paszke2017automatic}.

\begin{figure}[!bt]
  \centering
  \scalebox{0.6}{
  \includegraphics{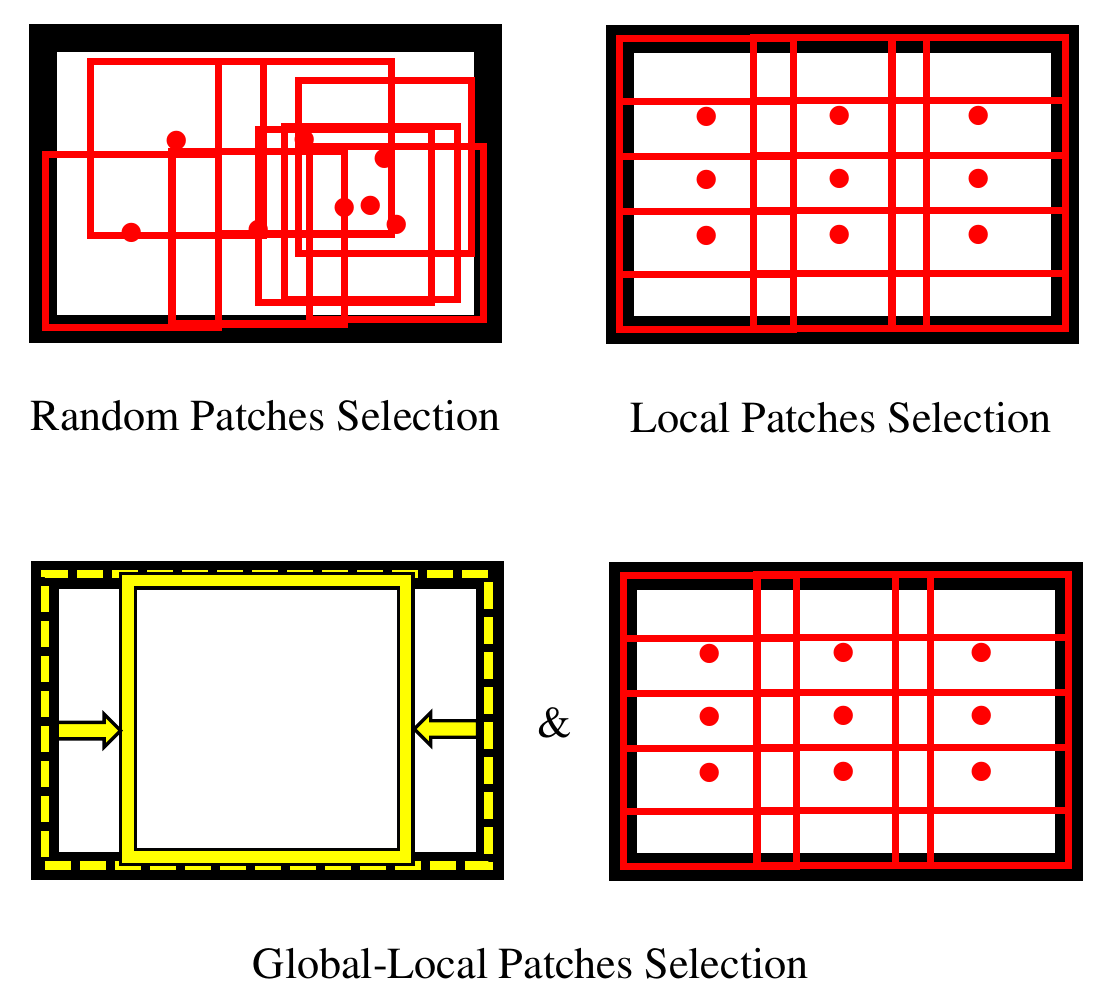}
  }
  \caption{Concepts of the three test patch-selection strategies.}
  \label{fig:test_patch_selection}
\end{figure}

\vspace{2mm}\noindent\textbf{Test. \ \ } \label{sec:ablation_patch_strategies}
The global-local multi-patch evaluation strategy (MP-GlobalLocal) described in \Sref{sec:test_flow} was compared with two other patch-selection strategies: one involving the use of local patches only (MP-Local) and one in which patches are cropped randomly (MP-Random). Examples of cropping process for these strategies are presented in \Fref{fig:test_patch_selection}. To compare the performance of these strategies, we tested cropping several numbers of patches.
In the MP-GlobalLocal strategy, we cropped one (one patch in a side), four (two patches in a side), and nine (three patches in a side) local $299 \times 299$ patches at equal intervals and resized the entire image to a global $342 \times 342$ patch. In addition, the same numbers of local patches were cropped in the MP-Local strategy, and one to ten randomly cropped $299 \times 299$ patches were used for the Random strategy.

\section{Results} \label{sec:results}
First, we compare the results of aesthetics score prediction performance obtained by changing the loss functions, patch-selection strategies, and number of test patches. 
Following that, we present an overall performance comparison with previous works using several metrics. Finally, we highlight the performance improvements obtained for each aspect ratio of images via aspect-ratio-preserving learning.

\subsection{Ablation Studies} \label{sec:ablation_result}
We employed the linear correlation coefficient (LCC) and root mean squared error (RMSE) to compare the aesthetics score prediction performance of our methods, including those of the ablation experiments. Eight loss functions and test patch-selection strategies described in \Sref{sec:ablation_training} were used. We compared the metrics of these models at the epochs when the best performance was achieved for the validation dataset. The results are shown in \Tref{tab:result_ablation_studies}.

\begin{table*}[!tb]
  \caption{Comparison of the aesthetics score prediction performance of the proposed \INDEMDstr-MP-GlobalLocal method and those of previous methods. The first nine rows present the results of previous methods and the bottom two rows indicate the results of our experiments. For each metric, the best value is shown in bold.
  }
  \begin{center}
  \begin{tabular}{llllll}
\hline\hline
\tabcenter{Models} & \tabcenter{LCC $\uparrow$} & \tabcenter{SRCC $\uparrow$} & \tabcenter{MSE $\downarrow$} & \tabcenter{acc [\%] $\uparrow$ }  & \tabcenter{EMD $\downarrow$} \\
\hline
GIST linear-SVR~\cite{kao2015visual}  & - & - & 0.0522 & - & -\\
GIST RBF-SVR~\cite{kao2015visual} & - & - & 0.5307 & - & - \\
BOV-SIFT linear-SVR~\cite{kao2015visual} & - & - & 0.5401 & - & - \\
BOV-SIFT RBF-SVR~\cite{kao2015visual} & - & - & 0.5513 & - & - \\
Kao \etal~\cite{kao2015visual} & - & -  & 0.4510 & - & - \\
Jin \etal~\cite{jin2016image} & - & -  & 0.3373 & - & - \\
Roy \etal~\cite{roy2018predicting} &  - & - & 0.3562 & - & - \\
NIMA (Inception-V2) rept. 2018~\cite{talebi2018nima} & 0.636 & 0.612 & - & 81.51 & 0.050 \\
GPF-CNN (InceptionNet)~\cite{zhang2019gated} & 0.7042 & 0.6900 & 0.2752 & \textbf{81.81} & \textbf{0.045} \\
\hline
NIMA (our impl. using Inception-V3) & 0.6914 & 0.6802 & 0.2830 & 79.88 & 0.066 \\
MPA-Net (\INDEMDstr-MP-GlobalLocal) (proposed) & \textbf{0.7096} & \textbf{0.7004} & \textbf{0.2637} & 80.09 & 0.064\\
\hline\hline
\end{tabular}
\end{center}
\label{tab:exp_result_predict}
\vspace{-10pt}
\end{table*}

As for the loss functions, \Tref{tab:result_ablation_studies} demonstrates that \textit{individual} asynchronous learning outperforms \textit{collective} simultaneous learning for every combination of the use of the logarithm of $\EMDc$ and the use of the weight coefficient $\omega_\beta$. Moreover, prediction performance improved in all cases in which the weight coefficient was enabled. However, no obvious relationship was observed between the aesthetics score prediction performance and the use of logarithmic $\EMDc$. 
In these experiments, the best aesthetics score prediction performance was achieved when training with $\INDEMD{}$, which uses the logarithm of $\EMDc$ and weight coefficient $\omega_\beta$. This holds for all patch-selection strategies.

\begin{figure}[bt]
  \centering
  \scalebox{0.5}{
  \includegraphics{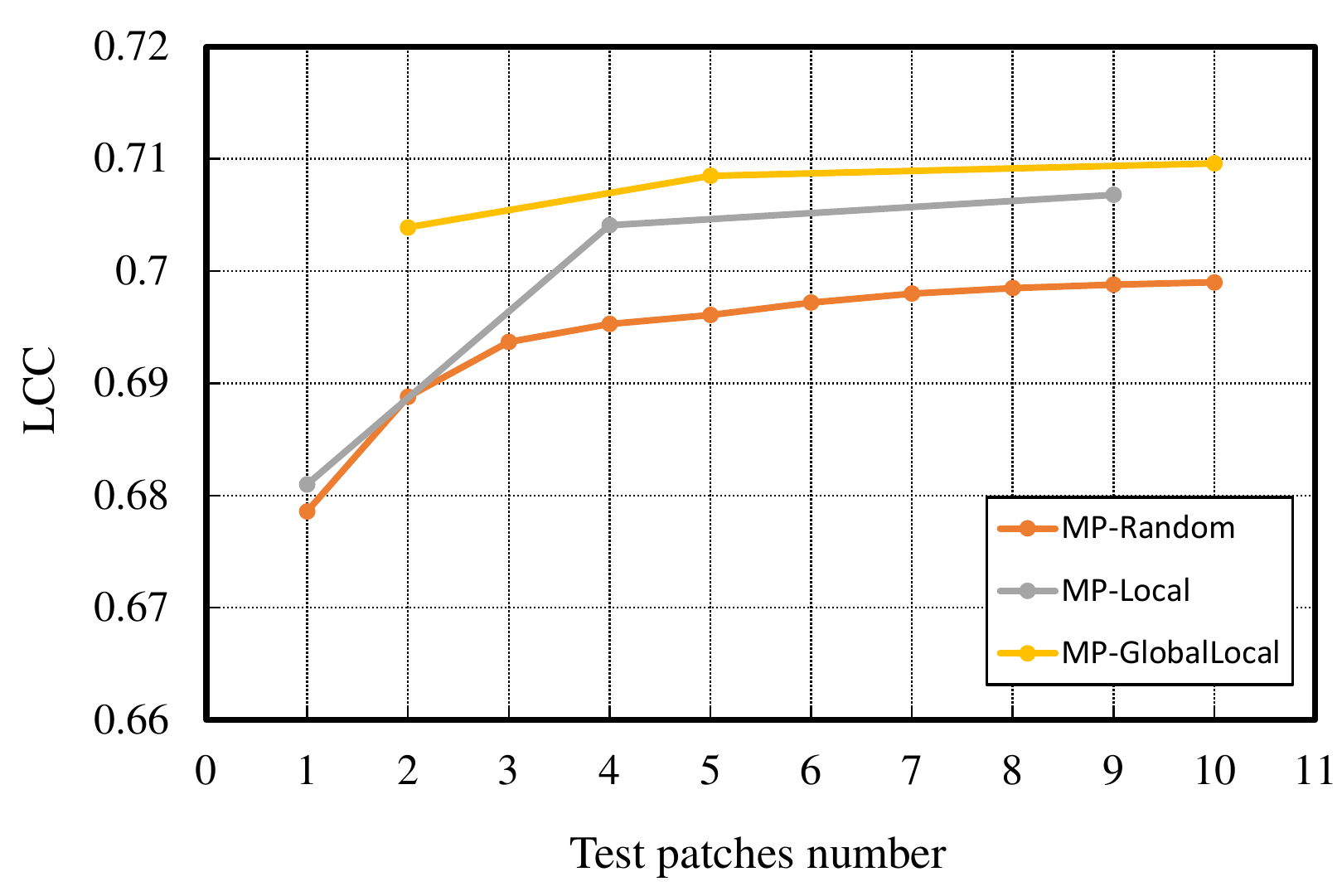}
  }
  \caption{LCC of the predictions of the $\INDEMD{}$ model versus the number of test patches for different patch-selection strategies (MP-Random, MP-Local, and MP-GlobalLocal).} 
  \label{fig:patch_num_LCC}
\end{figure}

\begin{figure}[bt]
  \centering
  \scalebox{0.5}{
  \includegraphics{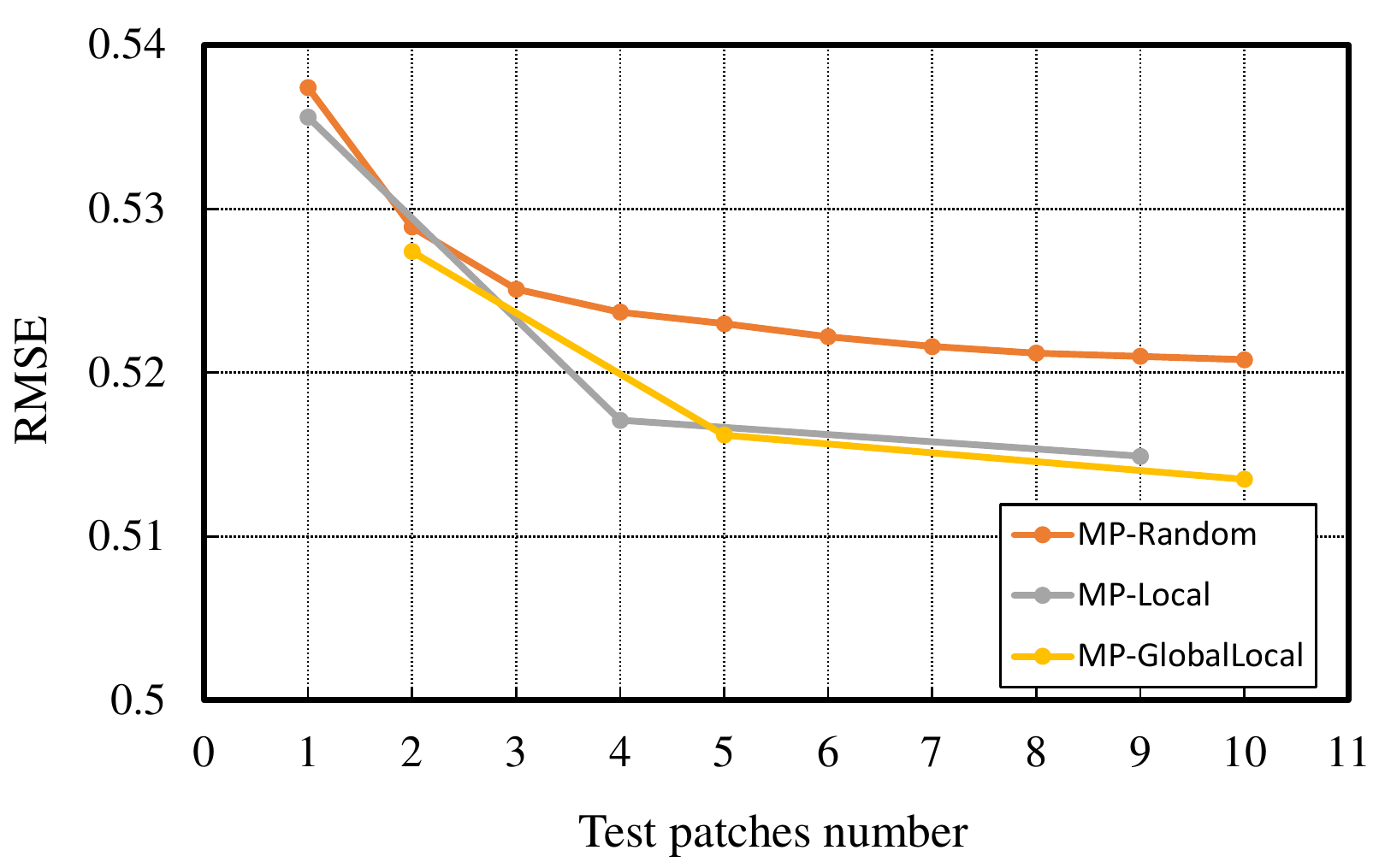}
  }
  \caption{RMSE of the predictions of the $\INDEMD{}$ model versus the number of test patches for different patch-selection strategies (MP-Random, MP-Local, and MP-GlobalLocal).} 
  \label{fig:patch_num_RMSE}
\end{figure}

As for test patches selection, we made a further detailed investigation on the selection strategies and the number of selected patches using the model trained with $\INDEMD{}$. The three patch-selection strategies, namely MP-Random, MP-Local and MP-GlobalLocal, were tested. The number of patches examined were 1 to 10 for MP-Random, {1, 4, 9} for MP-Local, and {2, 5, 10} for MP-GlobalLocal. The patch numbers for MP-Local and MP-GlobalLocal corresponded to the side patch numbers {1, 2, 3}.
\Fref{fig:patch_num_LCC} shows that the LCC of the predictions varied with the number of test patches for the three patch-selection strategies. This demonstrates that as the number of patches used for testing increases, the LCC gradually increases. Furthermore, compared with the MP-Random strategy, MP-Local yields better results, and MP-GlobalLocal even outperformed MP-Local. This trend was also observed in the RMSE of the predicted scores. \Fref{fig:patch_num_RMSE} shows the RMSE changes against the number of test patches. The RMSE became smaller as more test patches were used, and the MP-GlobalLocal strategy tended to show better performance than either MP-Random and MP-Local. 

Therefore, the model trained with the loss function $\INDEMD{}$ performed the best when using the MP-GlobalLocal test patch-selection strategy.

\subsection{Comparison with Existing Methods}
We compared the performance of aesthetics score prediction performance of methods proposed in previous works and {\INDEMDstr}-MP-GlobalLocal, which was the best of our models according to the experimental results. 
In addition to the LCC metric used in the ablation studies, we employed Spearman's rank correlation coefficient (SRCC) and mean squared error (MSE) for evaluating the prediction performance of the methods. Moreover, we calculated the accuracy (acc) of the models for aesthetics binary classification and the average EMD for comparison with NIMA~\cite{talebi2018nima}. For binary classification, images with aesthetics scores less than or equal to 5 were labeled as negative, and the rest were labeled as positive. Nonetheless, it should be kept in mind that the main purpose of the models presented in this study is aesthetics score prediction.

The results are shown in \Tref{tab:exp_result_predict}. 
The MPA-Net trained with the $\INDEMD{}$ loss function outperformed the baseline NIMA model reported by Talebi \etal~\cite{talebi2018nima} for all the common metrics evaluated for aesthetics score prediction; the LCC was 0.073 (11.5\%) higher and SRCC was 0.088 (14.4\%) higher. Furthermore, compared with all other previous methods, the proposed method achieved the best performance for aesthetics score prediction; the LCC was 0.0054 (0.77\%) higher, the SRCC was 0.0104 (1.44\%) higher, and the MSE was 0.0115 (4.18\%) lower compared with the GPF-CNN approach reported by Zhang \etal~\cite{zhang2019gated}, which achieved the best values out of all previous works. 

However, no improvement was observed in terms of the accuracy of aesthetic binary classification and the optimization of the EMD. The performance of the NIMA reported by Talebi \etal~\cite{talebi2018nima} and GPF-CNN~\cite{zhang2019gated} were superior to that of the proposed methods.

\begin{figure}[bt]
  \centering
  \scalebox{0.6}{
  \includegraphics{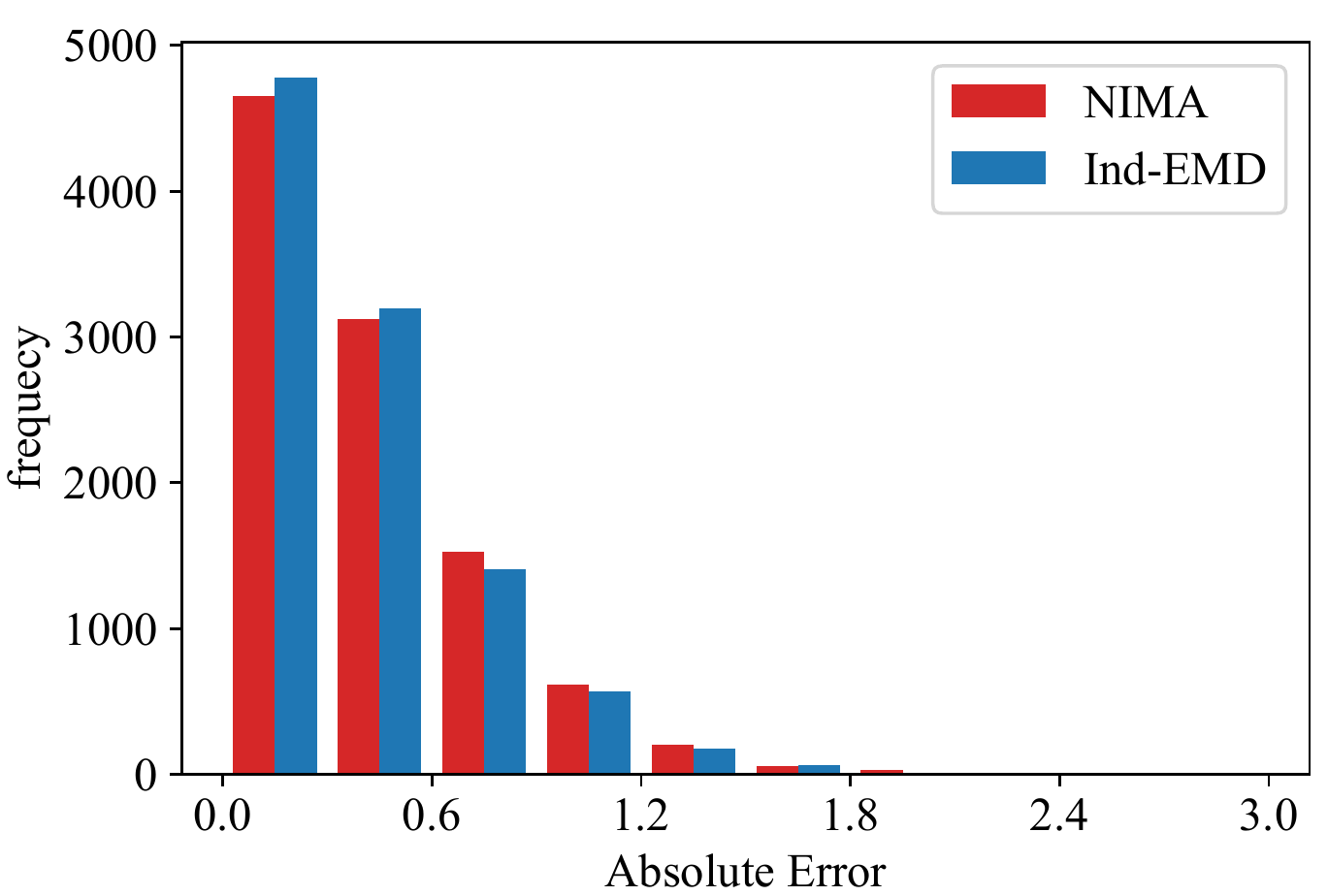}
  }
  \caption{Comparison of histograms of absolute errors of aesthetics scores predicted for the test dataset for the $\INDEMD{}$ model and the reproduced NIMA model.} 
  \label{fig:ada_ae_histogram}
\end{figure}

As a reference, a comparison of the histograms of the absolute errors (AEs) of the scores predicted by the $\INDEMD{}$ model and our implementation of the NIMA model for the test dataset is shown in \Fref{fig:ada_ae_histogram}.
\Fref{fig:ada_ae_histogram} demonstrates that the predicted aesthetics scores contain their AEs within 0.3 for approximately 45\% of the test images and within 0.6 for more than 75\% of the test images. Furthermore, the number of predictions with small AEs made by the $\INDEMD{}$ was larger than that of the reproduced NIMA model. Fewer predictions with AEs equal or larger than 0.5 were made by the $\INDEMD{}$ model than the reproduced NIMA model. Therefore, \Fref{fig:ada_ae_histogram} indicates that the proposed $\INDEMD{}$ method results in smaller errors for aesthetics score prediction.

Examples of predictions are shown in Appendix.

\subsection{Dependence of MSE Improvement on Image Aspect Ratio} \label{sec:MSE_aspect_ratio_results}

\begin{figure}[bt]
  \centering
  \scalebox{0.5}{
  \includegraphics{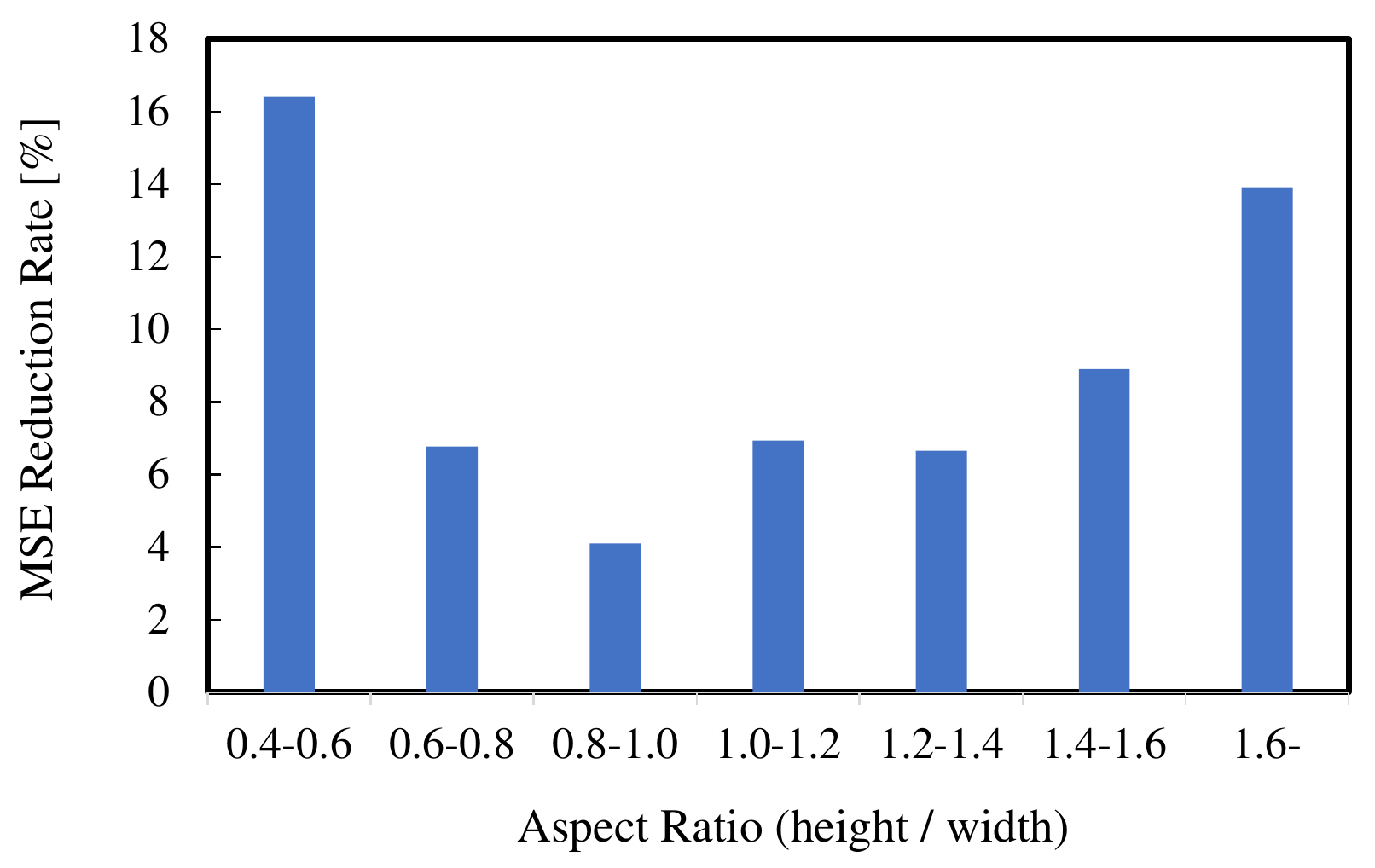}
  }
  \caption{MSE reduction rate for the aesthetics score predictions for various image aspect ratios of the model trained with $\INDEMD{}$ and the NIMA model.}
  \label{fig:aspect_improve}
\end{figure}

We also investigated the MSE improvement obtained by the proposed model trained with the $\INDEMD{}$ loss function for different image aspect ratios and compared it with that for the NIMA model. \Tref{tab:aspect_result} shows the MSE between the ground truth and the predictions for various height/width aspect ratios, and \Fref{fig:aspect_improve} shows the percentages of MSE reduction for different aspect ratios of images. From these results, it can be seen that the MSEs for images with aspect ratios within the ranges of 0.4--0.6 and 1.6-- were more likely to be reduced than for images with aspect ratios near 1.0 (0.8--1.0 and 1.0--1.2) or aspect ratios frequent in the training dataset, as described in \Sref{sec:dataset} (0.6--0.8, 1.2--1.4 and 1.4--1.6). In particular, the proportion of MSE reduction was at most 4.0 times larger for aspect ratios of 0.4--0.6 compared with aspect ratios of 0.8--1.0.
This can be ascribed to the ability of our model to fully use the information of the images over all areas while maintaining the original aspect ratios of contents, in contrast to the NIMA model, which deforms objects in the images by resizing them. Because NIMA does not preserve aspect ratios of contents, it tends to fit images with common aspect ratios or square images, and therefore does not work well for extraordinary aspect ratios far from 1.0. Our method mainly reduced the errors caused by this modification of the aspect ratios of contents and made it possible to manage a wide range of image aspect ratios.

\begin{table}[!tb]
  \caption{MSE of the predictions of the reproduced NIMA model and the proposed MPA-Net (\INDEMDstr-MP-GlobalLocal) model for various ranges of image aspect ratios.
  }
  \begin{center}
  \begin{tabular}{lll}
\hline\hline
\tabcenter{Image Aspect ratio} 
& \multicolumn{2}{c}{Model} \\
\cline{2-3} 
\tabcenter{(height/width)} & NIMA & MPA-Net \\
\hline
0.4--0.6    & 0.3152 & 0.2635 \\
0.6--0.8    & 0.2849 & 0.2656 \\
0.8--1.0    & 0.2723 & 0.2611 \\
1.0--1.2    & 0.3108 & 0.2892 \\
1.2--1.4    & 0.2924 & 0.2729 \\
1.4--1.6    & 0.2419 & 0.2203 \\
1.6--       & 0.2917 & 0.2511 \\
\hline\hline
\end{tabular}
\end{center}
\label{tab:aspect_result}
\vspace{-10pt}
\end{table}

\subsection{Discussion}

In this part, we provide a fine-grained interpretation of the experimental results for the components of our model.

According to the results of our detailed ablation studies, we found that \textit{individual} model updates are more effective than \textit{collective} model updates, and prediction performance increases when more patches are used for prediction. A possible reason for this is that, because the cropped patches from the same image are likely to be similar to each other, they tend to emphasize common elements and cancel the characteristic features of each patch in the \textit{collective} model updates. \textit{Individual} patch training can suppress feature cancellation within the same image. It should be noted that this discussion only focuses on simultaneous updating of patches from the same image and is not applicable for mini-batch training, in which the model is simultaneously updated using patches from different images. 

As for the other components of the loss function, the weight coefficient $\omega_\beta$ of each patch improved prediction performance. This result implies that it is effective to relatively alter the intensity of the updates among patches according to the distance between the prediction and the ground truth. The logarithm of the EMD was adopted with the intention of smoothing the training process and it generally, but not always, worked as expected. However, more investigation is required to unravel the conditions under which the logarithm of the EMD yields strongly positive effects. As for the test patch-selection strategies, our experimental results indicate that the model performed better when patches were selected at equal intervals than when patches were cropped at random. This implies that it is more beneficial to thoroughly reflect the whole area of images. 

Moreover, our investigation of the MSE reduction achieved by the proposed method compared with the reimplemented baseline NIMA model showed that MSE decreased largely for image aspect ratios far from 1.0 for MPA-Net. As mentioned in \Sref{sec:MSE_aspect_ratio_results}, this suggests that much of the prediction error stems from aspect-ratio-altering resizing, and we believe that a large proportion of the error caused by resizing is eliminated when using MPA-Net. However, errors still remain, some of which are inevitable because human aesthetics are subjective. Moreover, some errors may be reduced by employing other meta-information, such as image targets, although this exceeds the scope of this paper.

It should be noted that the proposed methods did not work well for aesthetic binary classification and EMD optimization. The reason for the observed low performance in binary classification tasks is considered to be the prediction bias around the classification threshold. Because a slight prediction bias near the classification threshold can largely affect classification accuracy, this result does not conflict with the success of aesthetics score predictions.
Generally speaking, score prediction is harder to optimize than binary classification. Thus, minimizing a loss for score prediction do not always fully optimize binary classification.
Besides, failing to optimize the EMD is also not incompatible with making successful aesthetics score predictions because we minimize variants of EMD, not EMD itself, at training.

\section{Conclusions}

We proposed methods of an aspect-ratio-preserving multi-patch aesthetics score prediction method, named MPA-Net. Through experiments using the AVA dataset~\cite{murray2012ava}, MPA-Net performed predictions with an LCC 0.073 (11.5\%) higher and SRCC 0.088 (14.4\%) higher compared with our baseline, NIMA~\cite{talebi2018nima}. Compared with the state-of-the-art method of continuous aesthetics score prediction, GPF-CNN~\cite{zhang2019gated}, the proposed method yielded an MSE 0.0115 (4.18\%) lower and achieved comparable performance in terms of LCC and SRCC. In particular, our model can achieve lower MSE in predictions for images with aspect ratios far from 1.0, of which there are relatively few samples in the dataset and undergo serious deformations via square resizing. This result indicates that MPA-Net can predict aesthetics scores accurately for a wide range of image aspect ratios. Our ablation studies also revealed that the equal-interval test patch-selection strategy was more effective than the random patch-selection strategy.

With the improvement of the preformance of aesthetics score prediction, MPA-Net can expand the way of practical applications of quantitative aesthetics evaluations, such as image recommendation and photo selection. MPA-Net also could be easily applied to other datasets or other tasks because it does not require any external information in neither the training nor the prediction stages. For example, it should be possible to apply the proposed approach for other human subjectivity assessments.


\section*{Acknowledgment}

This research is partially supported by JST-CREST (JP-MJCR1686) and the Grants-in-Aid for Scientific Research Numbers JP18H03339 and JP19K20289 from JSPS.


%





\ifCLASSOPTIONcaptionsoff
  \newpage
\fi





\bibliographystyle{IEEEtran}
\bibliography{IEEEabrv,reference}

\vfill


\clearpage
\appendix[Examples of the predictions of aesthetics scores] \label{sec:image_score_examples}

In this section, we demonstrate some examples of the predictions of image aesthetics scores. A ground-truth score distribution, a NIMA-predicted score distribution, and an MPA-Net-predicted score distribution are given for each image.

\begin{figure}[!h]
    \centering
    \begin{minipage}{1\hsize}
        \centering
        \scalebox{0.46}{
        \includegraphics{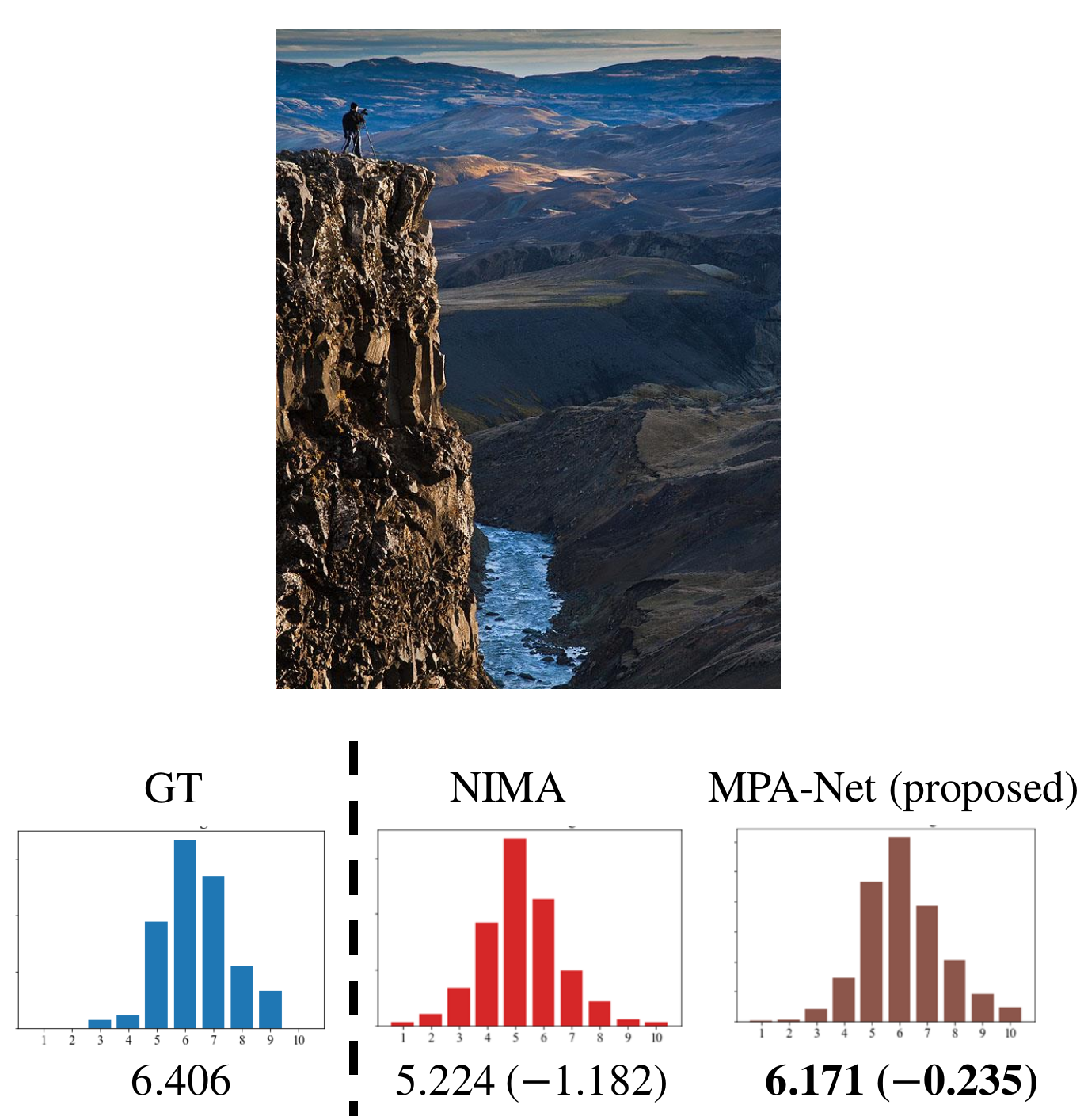}
        }
    \end{minipage}
    \begin{minipage}{1\hsize}
        \centering
        \scalebox{0.46}{
        \includegraphics{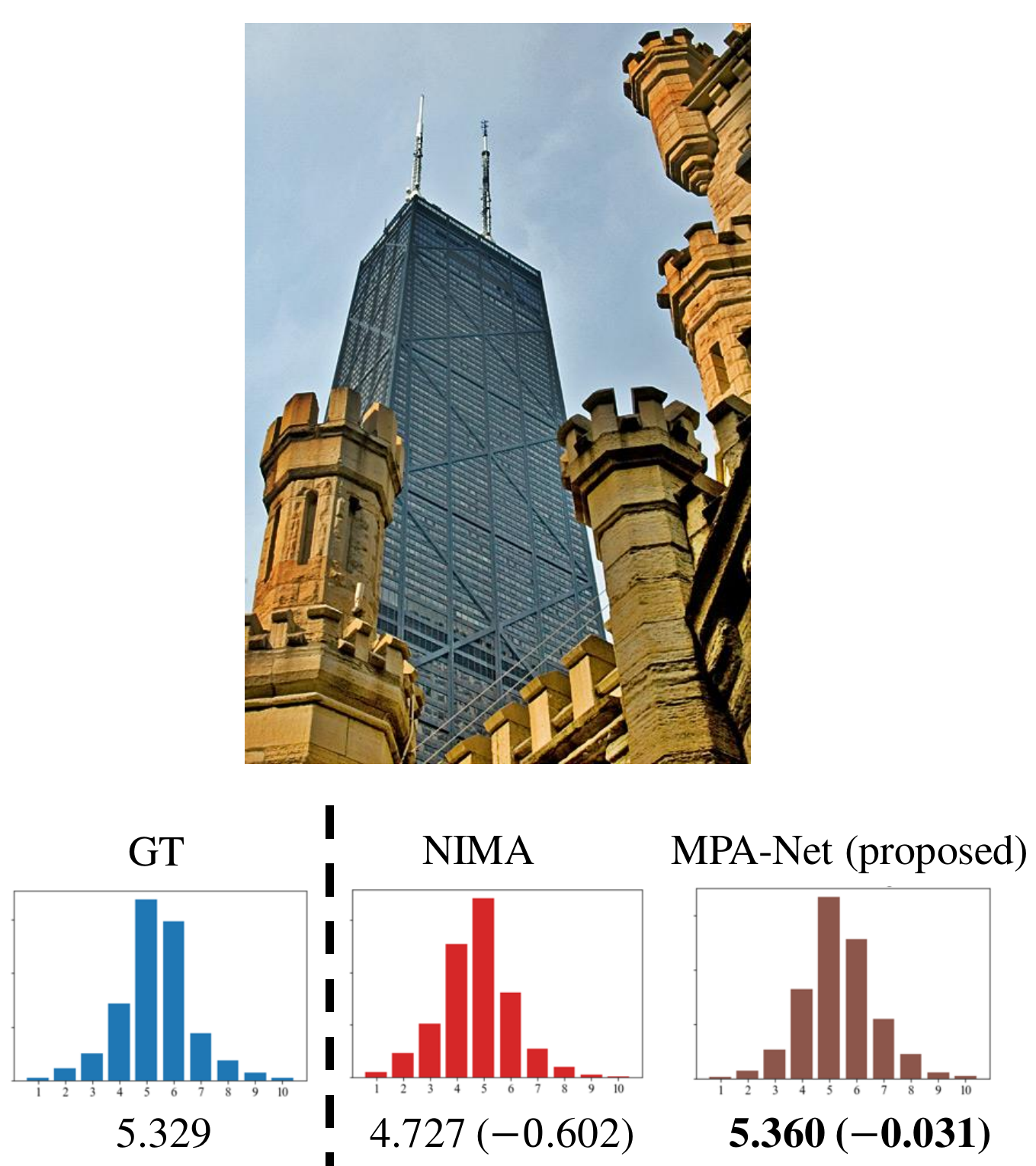}
        }
    \end{minipage}
    \begin{minipage}{1\hsize}
        \centering
        \scalebox{0.46}{
        \includegraphics{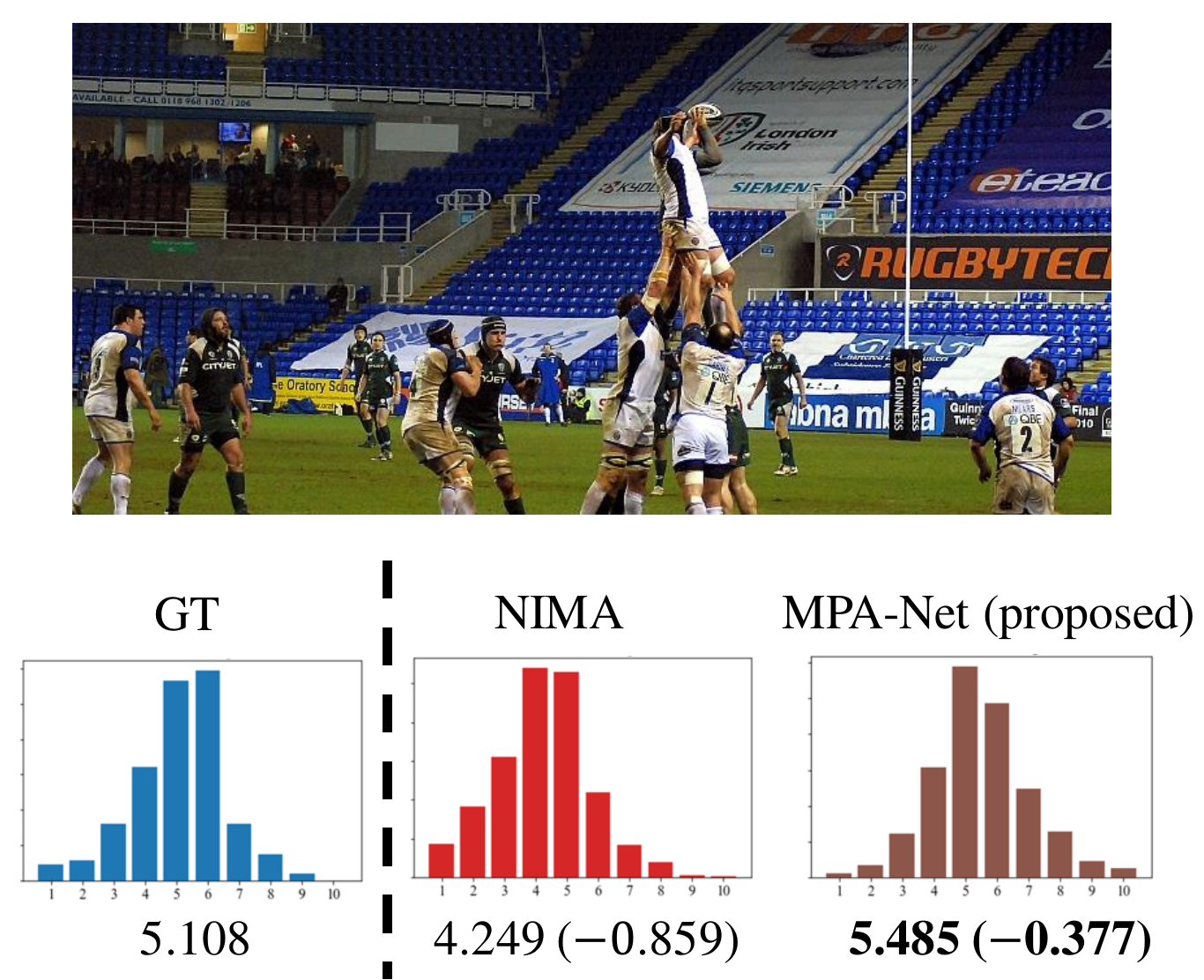}
        }
    \end{minipage}
    \caption{Examples of prediction improved by the MPA-Net model compared to the reproduced NIMA model. Numbers under distribution denote aesthetic scores and the number inside each bracket is the difference between the prediction and the ground truth.}
    \label{fig:predict_success}
\end{figure}

\begin{figure}[!htb]
    \centering
    \begin{minipage}{1\hsize}
        \centering
        \scalebox{0.46}{
        \includegraphics{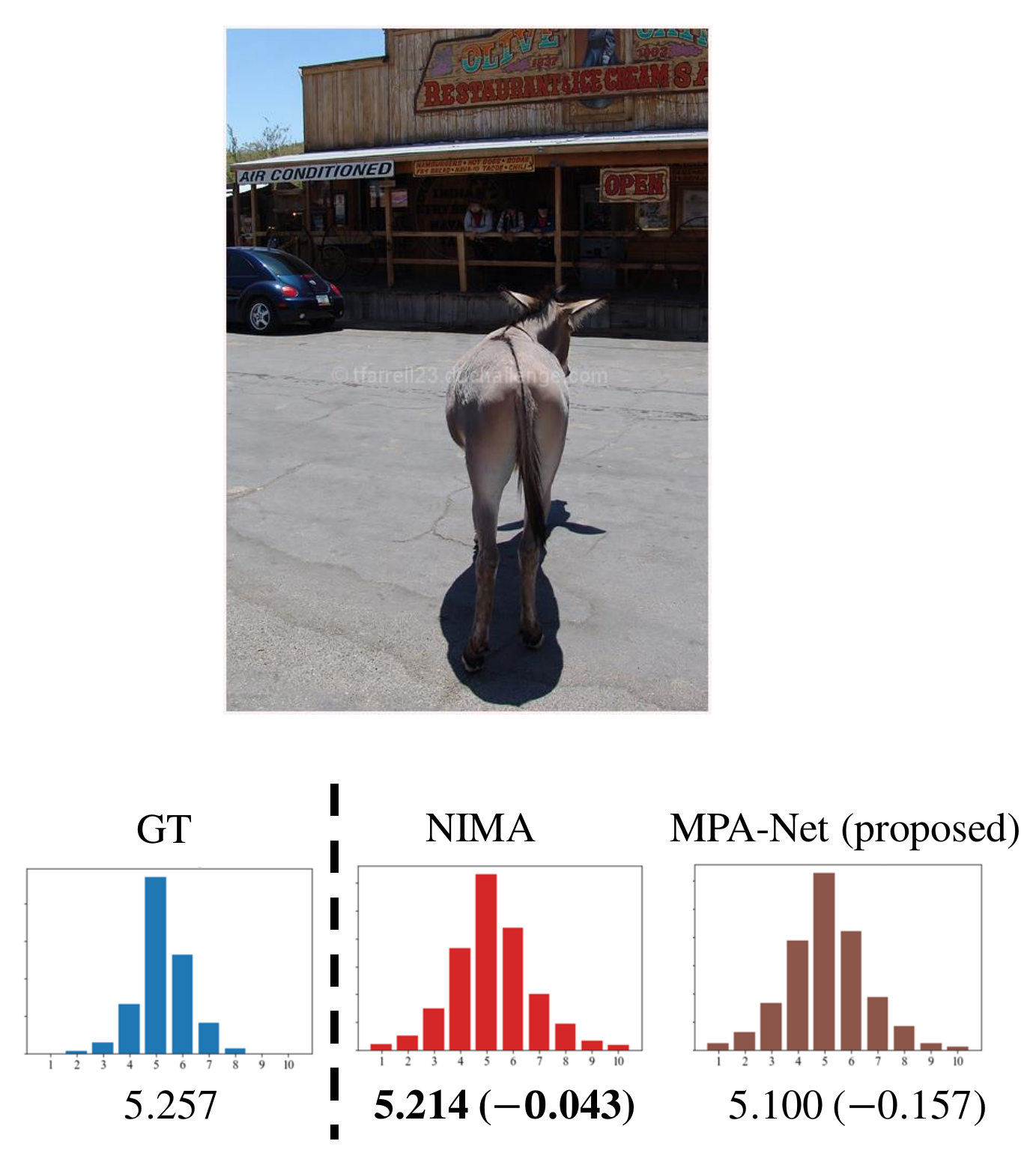}
        }
    \end{minipage}
    \begin{minipage}{1\hsize}
        \centering
        \scalebox{0.46}{
        \includegraphics{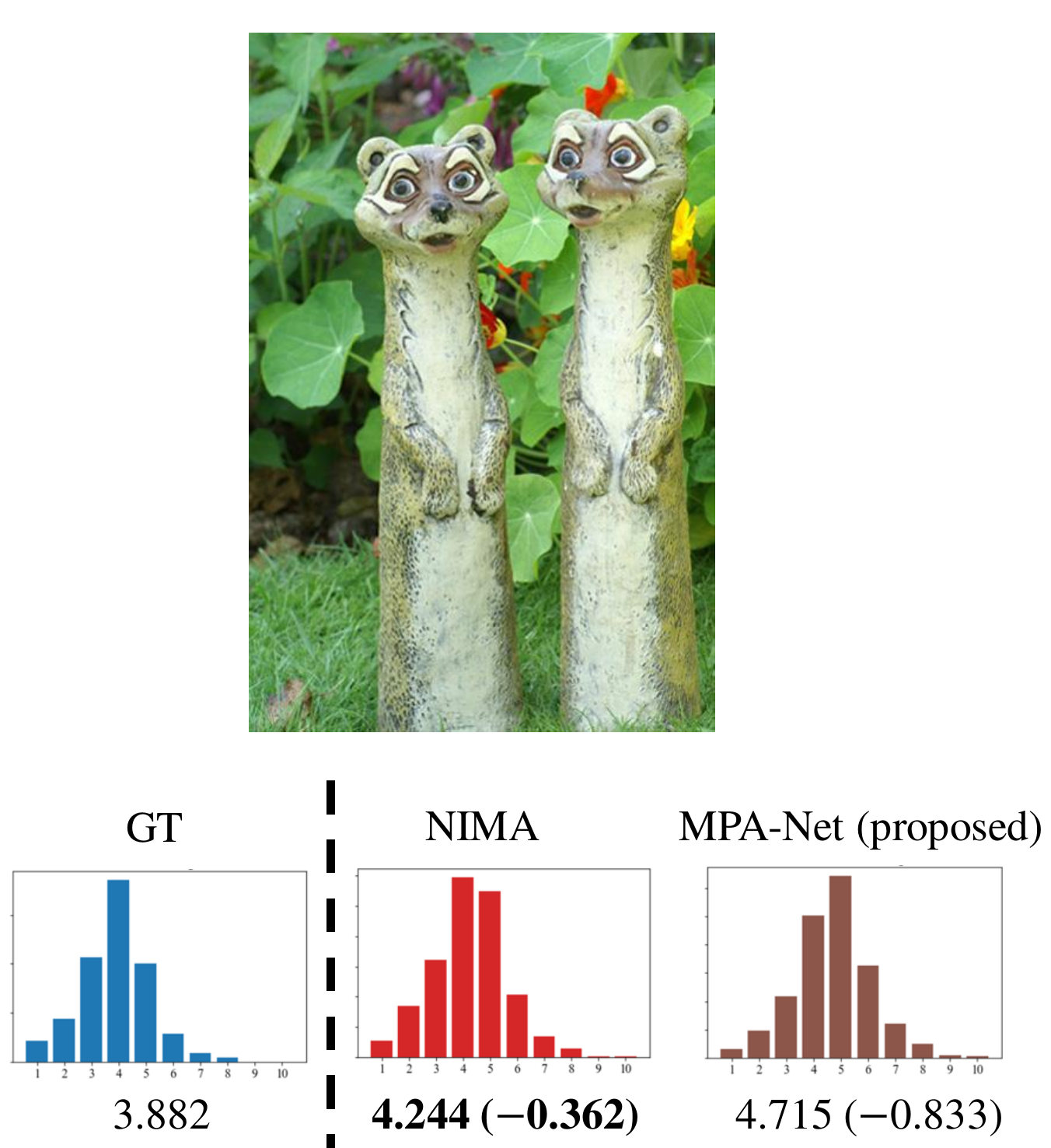}
        }
    \end{minipage}
    \caption{Examples of prediction deteriorated by the MPA-Net model compared to the reproduced NIMA model. Numbers under distribution denote aesthetic scores, and a number inside each bracket is the difference between the prediction and the ground truth.}
    \label{fig:predict_failure}
\end{figure}

Generally speaking, more predictions of aesthetics score were improved than that were deteriorated, and the degree of improvement was larger than the degree of deterioration.

\end{document}